\documentclass[journal]{IEEEtran}
\usepackage{amsmath,amsfonts}
\usepackage[ruled,vlined]{algorithm2e}
\usepackage{array}
\usepackage[caption=false,font=normalsize,labelfont=sf,textfont=sf]{subfig}
\usepackage{textcomp}
\usepackage{stfloats}
\usepackage{url}
\usepackage{verbatim}
\usepackage{graphicx}
\usepackage{cite}
\usepackage{booktabs}
\usepackage{floatrow}
\usepackage{indentfirst}
\usepackage{boldline}
\usepackage{multirow}
\usepackage{booktabs}
\usepackage{comment}
\usepackage{float}
\usepackage{tabularx}
\usepackage{makecell}
\usepackage{arydshln}
\usepackage{times}
\usepackage{epsfig}
\usepackage{graphicx}
\usepackage{amssymb}
\usepackage{algcompatible}
\usepackage{setspace}
\usepackage{soul,color}

\soulregister\cite7
\soulregister\ref7
\soulregister\pageref7

\makeatletter
\renewcommand{\ALG@beginalgorithmic}{\scriptsize \setstretch{1.1}} 
\makeatother

\usepackage{romannum}
\AtBeginDocument{\pagenumbering{arabic}}
\makeatletter

\let\saved@IEEEtitlepagestyle\ps@IEEEtitlepagestyle

\let\saved@@oddhead\@oddhead
\let\saved@@evenhead\@evenhead
\let\saved@@oddfoot\@oddfoot
\let\saved@@evenfoot\@evenfoot

\def\ps@IEEEtitlepagestylewithfooter{%
    
    \saved@IEEEtitlepagestyle

    \let\@oddhead\saved@@oddhead
    \let\@evenhead\saved@@evenhead

    \def\@oddfoot{%
      \hbox{}%
      \parbox{\textwidth}{%
        \centering
        \footnotesize
        
        Copyright © 2024 IEEE. Personal use of this material is permitted.
        However, permission to use this material for any other purposes must be
        obtained from the IEEE by sending an email to \texttt{pubs-permissions@ieee.org}.
      }%
      \hbox{}%
    }%
    \def\@evenfoot{\@oddfoot}%
}
\makeatother

\begin{document}

\title{MSCoTDet: Language-driven Multi-modal Fusion for Improved Multispectral Pedestrian Detection}

\author{Taeheon Kim$^*$, Sangyun Chung$^*$, Damin Yeom, Youngjoon Yu, Hak Gu Kim and Yong Man Ro,~\IEEEmembership{Senior Member,~IEEE.}\thanks{$^*$: Both authors contributed equally to this manuscript.\\ This work was conducted by Center for Applied Research in Artificial Intelligence (CARAI) grant funded by DAPA and ADD (UD230017TD) (\textit{Corresponding author: Yong Man Ro}).\\
T. Kim, S. Chung, D. Yeom, Y. Yu, and Y. M. Ro are with Integrated Vision and Language Lab., School of Electrical Engineering, Korea Advanced Institute of Science and Technology (KAIST), 291 Daehak-ro, Yuseong-gu, Daejeon, 34141, Republic of Korea (e-mail: eetaekim@kaist.ac.kr; jelarum@kaist.ac.kr; damin321@kaist.ac.kr; greatday@kaist.ac.kr; ymro@kaist.ac.kr).\\
H.G. Kim is with Department of Image Science and Arts, GSAIM, Chung-Ang University, Seoul, 06974, Republic of Korea (e-mail: hakgukim@cau.ac.kr).}}

\markboth{IEEE TRANSACTIONS on Circuits and Systems for Video Technology}%
{Shell \MakeLowercase{\textit{et al.}}: A Sample Article Using IEEEtran.cls for IEEE Journals}


\maketitle
\thispagestyle{IEEEtitlepagestylewithfooter}

\begin{abstract}
Multispectral pedestrian detection is attractive for around-the-clock applications due to the complementary information between RGB and thermal modalities. However, current models often fail to detect pedestrians in certain cases (e.g., thermal-obscured pedestrians), particularly due to the modality bias learned from statistically biased datasets. In this paper, we investigate how to mitigate modality bias in multispectral pedestrian detection using a Large Language Model (LLM). Accordingly, we design a Multispectral Chain-of-Thought (MSCoT) prompting strategy, which prompts the LLM to perform multispectral pedestrian detection. Moreover, we propose a novel Multispectral Chain-of-Thought Detection (MSCoTDet) framework that integrates MSCoT prompting into multispectral pedestrian detection. To this end, we design a Language-driven Multi-modal Fusion (LMF) strategy that enables fusing the outputs of MSCoT prompting with the detection results of vision-based multispectral pedestrian detection models. Extensive experiments validate that MSCoTDet effectively mitigates modality biases and improves multispectral pedestrian detection.
\end{abstract}


\section{Introduction}
\label{sec:intro}
\begin{figure*}[!t]
	\centering
	    \includegraphics[width=0.7\linewidth]{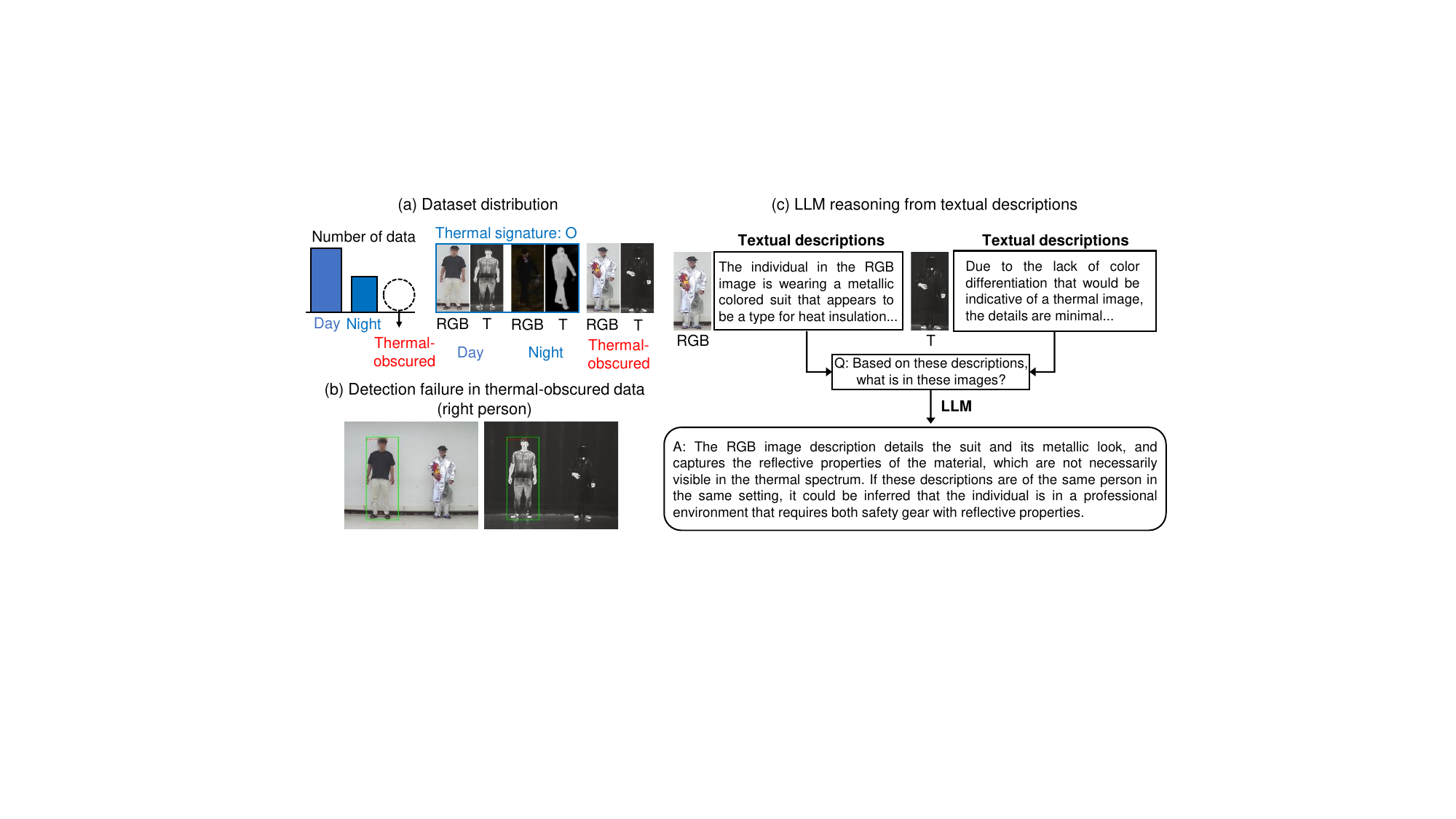}
	\caption{Problem illustration and our motivation. (a) In multispectral pedestrian datasets, thermal signatures always appear on pedestrians, as the thermal modality can generally capture pedestrians all day/night. In these datasets, thermal-obscured data is underrepresented. Models trained on such datasets learn the statistical co-occurrences between pedestrians and their thermal signatures. (b) As a result, models fail to detect pedestrians in thermal-obscured data, even though obviously visible in RGB. (c) An example of prompting the LLM. Based on the textual descriptions of RGB and thermal images, we prompt the ChatGPT~\cite{chatgpt} to answer the question ``\textit{Based on these descriptions, what is in these images?}". The ChatGPT detects the person without suffering modality biases, realizing that a person wearing heat-insulation clothing is invisible in thermal images due to the reflective material. Our motivation is to develop MSCoT prompting based on LLMs and integrate it into vision-based multispectral pedestrian detectors.}
\vspace{-0.3cm}
\label{unbalance}
\end{figure*}

\indent Multispectral pedestrian detection is the task of detecting pedestrians based on different visual modalities (i.e., RGB and thermal)~\cite{sun2022drone, guan2019fusion,kim2023multispectral, xu2017learning, liu2016multispectral, li2019illumination,zhang2019cross,zhang2019weakly, kim2022map,kim2024revisiting,kim2024mitigating}. Due to their complementary information, combining these modalities improves pedestrian detection all day/night~\cite{kim2021uncertainty, hwang2015multispectral, zhou2020improving,li2023stabilizing}. With advances in this field, the primary interest in multispectral pedestrian detection has been focused on effectively fusing the complementary information between the two modalities. Previous works investigated various fusion methods in different stages of the network, which are often categorized as early-fusion~\cite{wagner2016multispectral}, mid-fusion~\cite{kim2021uncertainty,kim2024causal, qingyun2022crossmodality}, and late-fusion~\cite{wagner2016multispectral, chen2022multimodal}. These methods demonstrated superior detection performance compared to standard pedestrian detection~\cite{kim2022defending,dollar2011pedestrian,hsu2020ratio,hasan2021generalizable,fan2023parformer,park2024integrating,shi2023fixated,bilal2016low,jeong2016early,chen2020high,kim2019bbc}, especially in practical datasets~\cite{c:25, hwang2015multispectral, gonzalez2016pedestrian} that contain all day/night scenarios. \\
\indent Despite the progress in this task, there are still remaining problems that need to be solved. From a recent work~\cite{kim2024causal}, multispectral pedestrian detection models are known to rely on spurious modality bias toward the thermal modality, due to learning the statistical bias in datasets. In multispectral pedestrian datasets, the pedestrian always statistically co-occur with its thermal signatures~\cite{kim2024causal} because the thermal modality is generally robust both day/night (Fig. \ref{unbalance} (a)). Models trained on these datasets learn the statistical co-occurrences of thermal signatures associated with pedestrians. As a result, models often fail on test data in which such co-occurrences do not hold, e.g., pedestrians with obscured thermal signatures (or, \textit{thermal-obscured} pedestrians). In the real world, thermal-obscured pedestrians are captured when intermediate obstacles, such as heat-insulating cloth or glass windows, block thermal radiation from reaching the thermal camera. Fig. \ref{unbalance} (b) demonstrates a multispectral pedestrian detector predicts the thermal-obscured pedestrian based on its absence of thermal signature, leading to failed detections. Such a phenomenon implies that modality biases limit the generalization of multispectral pedestrian detection models.\\
\indent However, developing a model that performs unbiased inference is challenging from biased training. This challenge arises due to a lack of explicit visual priors in training data. While increasing supervision of these priors through extra annotations or data augmentation could be one solution, it is difficult to achieve. It is because detection failures due to modality biases are not just limited to specific cases, such as thermal-obscured pedestrians, but are common in real-world data. For example, models often mistakenly detect trees or fireplugs as pedestrians due to their similar thermal signatures. It is difficult to find all detection failure cases caused by modality biases and augment the training priors accordingly~\cite{niu2021counterfactual, kim2024causal}. And the cost of multispectral data collection and manual annotation is prohibitive~\cite{chen2022multimodal}. \\  
\indent Our work is motivated by the observation that prompting large language models (LLMs)~\cite{kasneci2023chatgpt, achiam2023gpt, liu2024visual} can produce accurate detection results on challenging data (e.g., thermal-obscured pedestrians), especially in cases where conventional detection models fail due to modality biases. As shown in Fig.1 (c), we prompt an LLM with text descriptions of both RGB and thermal images, asking, “Based on these descriptions, what is in these images?”. From these descriptions, the LLM successfully detects the thermal-obscured pedestrian based on the rationale that heat-insulating clothing makes the pedestrian invisible in thermal images. In contrast, conventional vision-based detectors often fail to detect thermal-obscured pedestrians (Fig.1 (b)). These observations motivate us to adapt LLMs' rationale-based inference mechanisms for multispectral pedestrian detection. Specifically, by generating detection scores grounded in physical rationales through Chain-of-Thought prompting\mbox{~\cite{wei2022chain}}, we can integrate these scores into multispectral pedestrian detectors to help mitigate modality biases and enhance detection accuracy.  \\
\setlength{\textfloatsep}{10pt plus 1.0pt minus 2.0pt}
\sloppy \indent To this end, we design the Multispectral Chain-of-Thought (MSCoT) prompting, which prompts the LLM to perform the task of multispectral pedestrian detection. For the inputs of MSCoT prompting, we obtain textual descriptions of pedestrians in both RGB and thermal images from a pre-trained multi-modal language model (MLLMs)~\cite{achiam2023gpt}. Our goal of MSCoT prompting is to produce probabilistic detection scores from these textual descriptions. However, LLMs are known for their overconfidence~\cite{xiong2023can,li2024think,detommaso2024multicalibration} and output confidence scores irrelevant to the context~\cite{wightman2023strength}. To overcome this problem, we designed MSCoT prompting to conduct two consecutive Chain-of-Thought~\cite{wei2022chain} (CoT) prompting steps. MSCoT prompting first outputs uni-modal confidence scores and comprehends them with the text of both modalities to produce final detection scores. From these processes, MSCoT prompting facilitates LLMs to produce reliable detection scores that are compatible with multispectral pedestrian detectors.   \\
\indent Building on our MSCoT prompting, we propose the MSCoTDet framework that produces unbiased detections from a statistically biased train set. This problem formulation is adopted rather than explicitly modifying the training priors (such as adding thermal-obscured pedestrians in the train set) to verify that MSCoT prompting can effectively intervene in the modality bias of multispectral pedestrian detectors. Specifically, the
MSCoTDet framework integrates the decision processes of MSCoT prompting into vision-based multispectral pedestrian detectors. A Language-driven Multi-modal Fusion (LMF) is designed to effectively integrate these two distinct decision processes. \\
\indent To evaluate the effectiveness of MSCoTDet, we conduct experiments on generic multispectral pedestrian datasets (FLIR~\cite{c:25}, and CVC-14~\cite{gonzalez2016pedestrian}) and a dataset (ROTX-MP~\cite{kim2024causal}) that mainly contains thermal-obscured pedestrians. Extensive experiment results show that MSCoTDet generalizes well on ROTX-MP even with biased training data while performing robustly on generic datasets.\\
\textbf{Contributions.} In summary, our contributions are:
\begin{enumerate}
  \item We propose the Multispectral Chain-of-Thought (MSCoT) prompting strategy, which prompts the LLM to perform the task of multispectral pedestrian detection.
  \item We propose the MSCoTDet framework, which integrates MSCoT prompting into vision-based multispectral pedestrian detectors. 
  \item Extensive experiments demonstrate that our proposed MSCoTDet framework can significantly intervene in modality bias and improve the overall performance of multispectral pedestrian detectors. 
\end{enumerate}
\vspace{-0.3cm}
\section{Related Work}
\subsection{Multispectral Pedestrian Detection}
Multispectral pedestrian detection is the task of locating pedestrians in the input image based on multiple visual modalities (i.e., RGB and thermal). It is different from multispectral person re-identification~\cite{pu2020dual} which focuses on matching the same identity across a gallery of images to the query image of the opponent modality. Multispectral person re-identification lacks the detection process, and multispectral pedestrian detection does not consider identity-matching. Moreover, multispectral pedestrian detection stands apart from the task of person search~\cite{han2019re} which combines both detection and re-identification, aiming to simultaneously locate and identify a person from input images. \\
\textbf{Dataset.} Different from standard pedestrian detection~\cite{cao2019high, kim2020cua,zhang2020refinedet++,ouyang2015partial,jiao2020pen, park2024robust,yu2022defending}, multispectral pedestrian detection aims to detect pedestrians using RGB and thermal images. One challenge in multispectral pedestrian detection is the lack of data. The process of aligning RGB and thermal images requires expensive hardware, such as a beam-splitter~\cite{hwang2015multispectral} and a GPS clock. For annotating bounding boxes of multispectral data, each modality needs separate annotations.\\
\indent The KAIST~\cite{hwang2015multispectral} dataset is one of the first benchmarks for multispectral pedestrian detection, containing well-aligned RGB and thermal images. The FLIR~\cite{c:25} data offers higher image resolution than KAIST~\cite{hwang2015multispectral}. On the other hand, the CVC-14~\cite{gonzalez2016pedestrian} dataset largely contains misaligned RGB-thermal images. ROTX-MP~\cite{kim2024causal} dataset highlights the challenge due to the dataset bias of multispectral data. ROTX-MP adopts the evaluation setting where the test distribution is significantly different from the training priors. Performing unbiased
inference with biased training data remains a challenge in multispectral pedestrian detection. \\
\indent \textbf{Models.} Previous studies focused on developing various fusion methods for multispectral pedestrian detection. These approaches are based on early-fusion, mid-fusion, or late-fusion, which are often categorized by the fusion stage of the network. Early-fusion~\cite{wagner2016multispectral} methods concatenate RGB and thermal images at the input stage and processes through the detection network. Mid-fusion~\cite{wagner2016multispectral,liu2016multispectral,qingyun2022crossmodality,kim2021uncertainty} has been the prior focus, which encodes RGB images and thermal images to the same feature space and fuses them inside the network. However, early-fusion and mid-fusion are known to memorize the training priors and exhibit poor generalizability~\cite{kim2024causal} on out-of-distribution data. Lastly, late-fusion~\cite{chen2022multimodal,li2018multispectral} methods are based on ensembling detection results produced from RGB and thermal single-modal detectors. Most of them do heuristic weighting with single-modal detectors or perform simple Bayesian estimation, leading to reduced accuracy.\\
\begin{figure*}[!t]
	\centering
	    \includegraphics[width=0.75\linewidth]{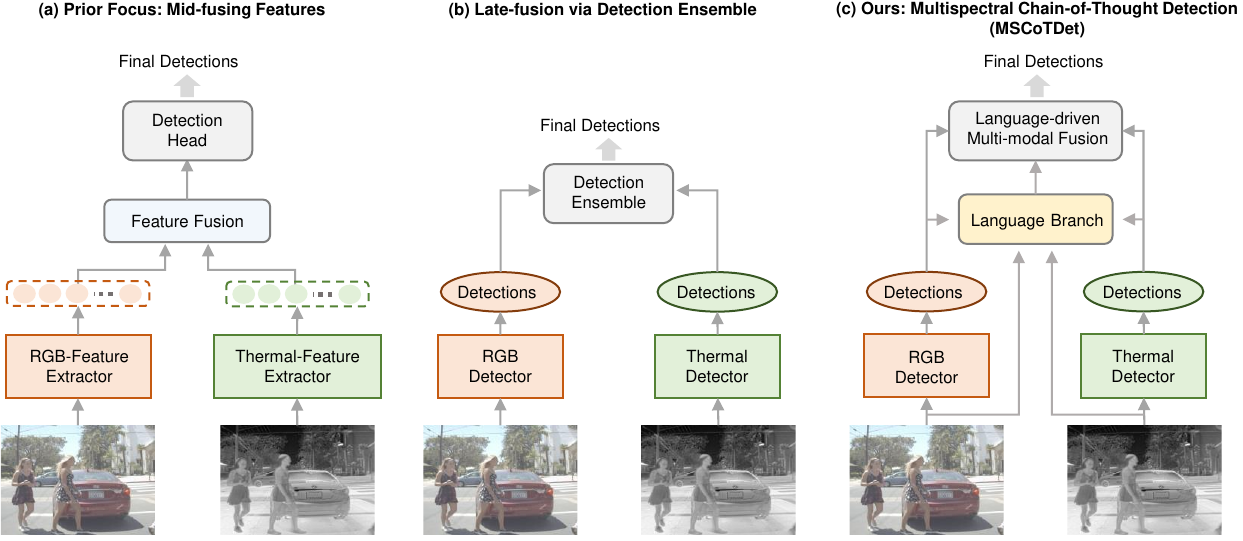}
	\caption{Comparisons between previous works and our method (MSCoTDet). (a) Previous approaches centrally focused on mid-fusion methods, e.g., mid-fusing features internally in the network. (b) There are few works via late fusion that ensemble detections from independently trained single-modal detectors, i.e., RGB and thermal detectors. (c) MSCoTDet (Ours) focuses on designing a language branch that processes detection using Large Language Models (LLMs). The language branch includes the MSCoT prompting, which prompts the LLM to perform multispectral pedestrian detection. Then, our proposed Language-driven Multi-modal Fusion (LMF) enables fusing vision-driven and language-driven detections.}
\end{figure*}
\indent Specifically, Kim et al.~\cite{kim2021uncertainty} devised an uncertainty-aware multispectral pedestrian detection framework that focuses on reducing the effect of the predictive uncertain modality (e.g., modality with corruption). The stable multispectral pedestrian detection (SMPD)~\cite{li2023stabilizing} model exploits the Dirichlet distribution to estimate the uncertainty distribution of different modalities and obtain fusion evidence by the Subjective Logic framework. Different from the uncertainty-based models~\cite{kim2021uncertainty, li2023stabilizing}, our work utilizes a large language model (LLM) to estimate the modal reliability. This is based on our observations that LLMs offer accurate modal reliability when the descriptions of the modalities are given as texts. ProbEn~\cite{chen2022multimodal} probabilistically ensembles detection results obtained from single-modal detectors of different modalities. As ProbEn's effectiveness depends on accurate probability distributions from both modalities, its effectiveness is degraded if one modality provides highly uncertain or miscalibrated predictions. MSCoTDet mitigates such problems by instructing the LLM to generate detection scores, which can make informed predictions based on the more reliable modality.\\
\indent MS-DETR~\cite{xing2024ms} expands the DETR architecture~\cite{zhu2020deformable} from a single-modal detection into the multispectral setting by devising a multi-modal transformer decoder. Compared to MS-DETR~\cite{xing2024ms}, MSCoTDet introduces a Language-driven Multi-modal Fusion strategy that combines single-modal DETR models with the inference mechanisms of a large language model (LLM). Lastly, the Causal Mode Multiplexer (CMM)\mbox{~\cite{kim2024causal}} framework aims to learn the causality between multispectral inputs and predictions using counterfactual interventions. However, such a method involves subtracting the direct effect from the model's detection score, resulting in low confidence levels. MSCoTDet facilitates LLMs to produce detection scores with robust confidence levels using Chain-of-Thought (CoT) Prompting~\cite{wei2022chain} mechanisms.\\
\vspace{-0.2cm}
\subsection{Addressing Modality Biases in Multimodal Learning} 
\indent As modality biases significantly influence the reliability of multi-modal models, several endeavors have been devoted to solving modality bias issues in different multi-modal tasks such as classification~\cite{wang2020makes}, visual question answering (VQA)~\cite{niu2021counterfactual}, and person re-identification~\cite{li2022counterfactual}. One straightforward approach to mitigate modality bias is to augment the training data or use additional annotations. In particular, counterfactual generation~\cite{abbasnejad2020counterfactual,liang2020learning,gokhale2020mutant,chen2020counterfactual} helps to balance the training data by explicitly adding the out-of-distribution priors. However, in the real world, unknown test cases are prevalent, and augmenting all these cases is difficult due to the vast diversity of real-world data~\cite{kim2024causal, niu2021counterfactual}. Therefore, various benchmarks (e.g., VQA-CP~\cite{agrawal2018don}, ROTX-MP~\cite{kim2024causal}) were proposed to evaluate whether multimodal models generalize well in subtly different distributions from training data.
\\
\indent Under these benchmarks, Wang et al.~\cite{wang2020makes} proposed Gradient Blending, which combines gradient estimates by calculating optimal weights for each modality. Gat et al.~\cite{gat2020removing} leverages the log-Sobolev inequality to bound the functional entropy, effectively maximizing the utilization of each modality's information during training. Niu et al.~\cite{niu2021counterfactual} capture the language bias in the VQA task as the natural direct effect of questions on answers. The debiasing is achieved by subtracting the language bias from the total effect. Similarly, our previous work~\cite{kim2024causal} aims to mitigate the thermal modality bias in multispectral pedestrian detection through counterfactual intervention. They subtract the detection score of the direct thermal path from the total detection score. Different from previous methods that regularize modality's gradients~\cite{wang2020makes, gat2020removing} or apply counterfactual intervention~\cite{niu2021counterfactual,kim2024causal}, our MSCotDet introduces an innovative approach utilizing language and chain of thought. MSCotDet mitigates modality bias by fusing modalities based on the semantic context between multimodal inputs and outputs using LLMs, making it robust to out-of-distribution data. \\
\vspace{-0.7cm}
\subsection{Vision Tasks with LLMs} 
Large language models (LLMs) excel in natural language processing (NLP) and generation by training on vast amounts of text data~\cite{ouyang2022training,brown2020language,workshop2022bloom,touvron2023llama,chowdhery2023palm,hoffmann2022training}. These features and advantages offer innovative applicability across more general areas of intelligence, such as computer vision. Yang et al.~\cite{yang2023language} introduced LLM-guided concept bottlenecks (LaBo) for image classification. LaBo generates candidate concepts from GPT-3~\cite{brown2020language}
and align them to test images by computing similarity scores. Naeem et al.~\cite{naeem2023i2mvformer} enhance zero-shot image classification framework utilizing an LLM to create web-scale documents. They compute the multiple complementary views of each class between text and image features and perform image classification. Khan et al.~\cite{khan2023q} improve VQA models by annotating unlabeled images with an LLM.\\
\indent Park et al.~\cite{park2024integrating} proposed a pedestrian detection framework that integrates LLM-derived text features via clustering and task-prompting. They create a document of pedestrian appearance descriptions using an LLM and combine them with visual features inside the pedestrian detection network. Although previous methods have enhanced model performance using LLMs, they 1) create large documents with LLMs\mbox{~\cite{brown2020language,naeem2023i2mvformer,park2024integrating}}, 2) combine/compare text and image features internally in the network\mbox{~\cite{brown2020language,naeem2023i2mvformer,park2024integrating}}, or 3) augment training data~\cite{khan2023q}. Different from previous works, we focus on prompting the LLM to perform the vision task (i.e., multispectral pedestrian detection) so that it fully exploits its linguistic capabilities. Furthermore, our proposed method completely sidesteps the need for constructing large documents or collecting/augmenting additional data.
\section{Preliminaries}
Before introducing our method, we review the late-fusion strategies in multispectral pedestrian detection. Then we elaborate on our proposed method.
\vspace{-0.2cm}
\begin{figure*}[t]
\centering
\includegraphics[width=0.9\textwidth]{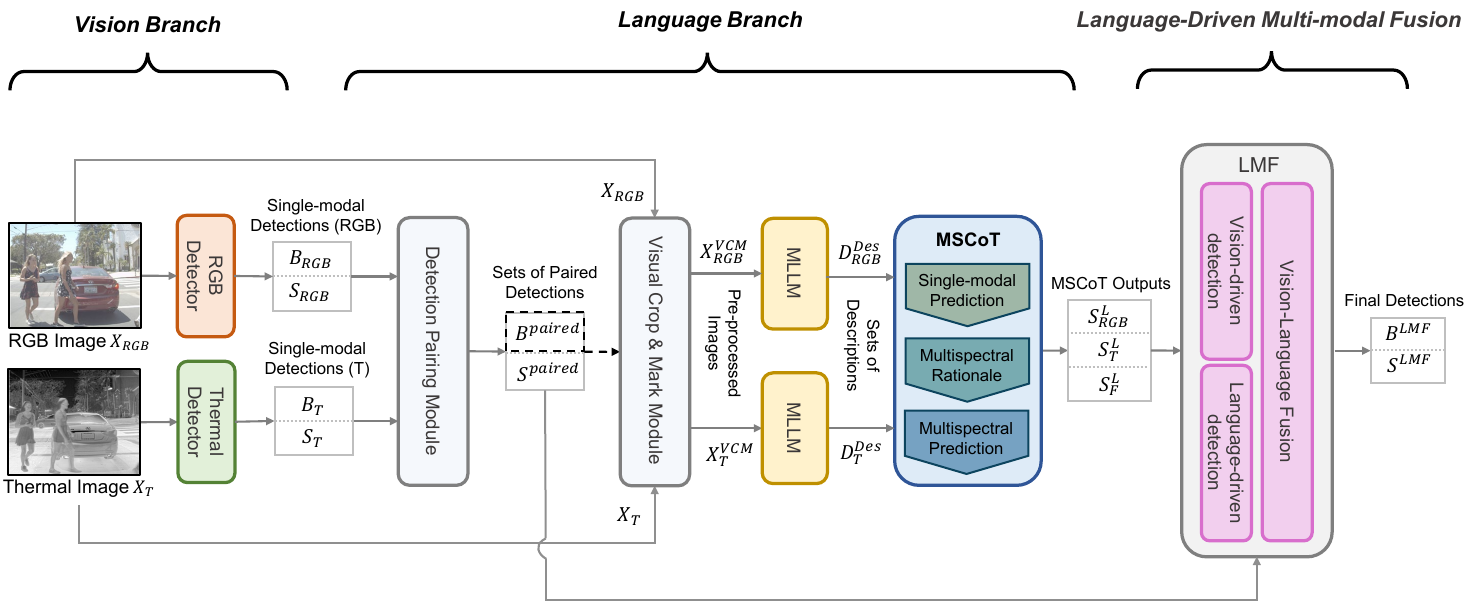}
\caption{Overall architecture of proposed Multispectral Chain-of-Thought detection (MSCoTDet) framework including vision branch, language branch, and Language-driven Multi-modal Fusion.}
\label{overall_arch}
\end{figure*}
\subsection{Late-fusion in Multispectral Pedestrian Detection} 
Rather than fusing modalities in the input stage (early-fusion~\cite{snoek2005early,tziafas2023early,nguyen2021gefa}) or in the intermediate steps (mid-fusion~\cite{kim2021uncertainty, qingyun2022crossmodality}), late-fusion~\cite{chen2022multimodal,chen2024s} is based on detection ensembling, i.e., fusing the decision values. One of the advantages of late fusion is that it can preserve single-modal models and, therefore, has high flexibility to perform multi-modal fusion from diverse models of different modalities~\cite{chen2024s,morvant2014majority, shutova2016black, yao2022modality, hessel2020does, pandeya2021deep}, \textit{e.g., fusing predictions between LLMs and vision-based detectors}.\\
Late fusion methods have been also investigated in multispectral pedestrian detection~\cite{chen2022multimodal}. They first train single-modal detectors (i.e., RGB and thermal) and then fuse the prediction scores (score fusion) and bounding boxes (box fusion) which are the outputs from single-modal detectors. The final output includes fused prediction scores and bounding boxes. We describe score fusion and box fusion strategies below.\\
\indent \textbf{Score Fusion.} The most common strategies to fuse prediction scores obtained from single-modal models are averaging~\cite{li2019illumination,liu2016multispectral}, Non-maximum Suppression (NMS)~\cite{chen2022multimodal}, and Probabilistic Ensembling (ProbEn~\cite{chen2022multimodal}). Between them, we adopt averaging and NMS, as we find them effective for our method (Section \ref{sec:3.4}). Averaging is straightforward, the fused score is determined by averaging prediction scores estimated in each modality. Denote $s_{\scriptscriptstyle RGB}$ and $s_{\scriptscriptstyle T}$ the prediction scores predicted for the same object in different modalities (i.e., RGB and thermal). Then the averaging score fusion between $s_{\scriptscriptstyle RGB}$ and $s_{\scriptscriptstyle T}$ can be expressed as: 
\begin{equation}{
s_{\scriptscriptstyle AVG}= \frac{s_{\scriptscriptstyle RGB}+s_{\scriptscriptstyle T}}{2},
\label{eq:1}
}\end{equation}
where $s_{\scriptscriptstyle AVG}$ denotes the fused score. On the other hand, NMS compares the prediction scores estimated in each modality and votes for the highest one, removing the lower scores. NMS score fusion between $s_{\scriptscriptstyle RGB}$ and $s_{\scriptscriptstyle T}$ can be expressed as: 
\begin{equation}{
s_{\scriptscriptstyle NMS}=max(s_{\scriptscriptstyle RGB}, s_{\scriptscriptstyle T}),
\label{eq:2}
}\end{equation}
where $s_{\scriptscriptstyle NMS}$ denotes the fused score by the NMS.
\\
\indent \textbf{Box Fusion.} Box fusion is to merge overlapping bounding boxes predicted from different modalities (i.e., RGB and thermal). Chen et al.(2022)\cite{chen2022multimodal} suggests a simple and effective way to probabilistically fuse boxes, which computes a weighted average of boxes. The weights are given by the prediction scores, implying that more confident detections should have a higher weight when fusing boxes. Denote $b_{\scriptscriptstyle RGB}$ and $b_{\scriptscriptstyle T}$ the bounding box coordinates predicted for the same object in different modalities (i.e., RGB and thermal). Using the predictions scores $s_{\scriptscriptstyle RGB}$, and $s_{\scriptscriptstyle T}$, the weighted-averaging box fusion between $b_{\scriptscriptstyle RGB}$ and $b_{\scriptscriptstyle T}$ can be expressed as:
\begin{equation}{
b_{\scriptscriptstyle s\text{-}avg }=\frac{b_{\scriptscriptstyle RGB} \times s_{\scriptscriptstyle RGB}+ b_{\scriptscriptstyle T} \times s_{\scriptscriptstyle T}}{s_{\scriptscriptstyle RGB} + s_{\scriptscriptstyle T}},
\label{eq:3}
}\end{equation}
where $b_{\scriptscriptstyle s\text{-}avg }$ denotes the fused bounding boxes.\\
\indent \textbf{Our Motivation.} Previous works have studied late-fusion approaches in multispectral pedestrian detection with two single-modal detectors, i.e., an RGB object detector and a thermal object detector. Most do heuristic weighting with single-modal detectors or perform simple Bayesian estimation~\cite{chen2022multimodal}. Therefore, these methods sometimes neglect the inter-modality contexts that might differ between the diverse scenarios of multispectral pedestrian detection. One advantage of late-fusion is that it preserves the single-modal predictions, as it does not require aligning different modalities in the same domain (early-fusion) or feature space (mid-fusion). Our motivation is to leverage the late-fusion strategies to fuse predictions from LLMs and vision-based detection models. To this end, we design the Language-driven Multi-modal Fusion (LMF) in Section \ref{sec:3.4}. In the next section, we propose the MSCoTDet framework and describe its details.   \\
\section{Proposed Method}
\subsection{Overall Architecture}
The overall architecture of our proposed MSCoTDet framework is shown in Fig. \ref{overall_arch}. It consists of two branches that perform distinct detection processes, the vision branch and the language branch. Detections from these branches are fused to output final detections. We briefly introduce the process of each branch and our fusion strategy.\\ 
\indent \textbf{Vision branch.} Given the input of RGB and thermal images $X_{\scriptscriptstyle RGB}$ and $X_{\scriptscriptstyle T}$, the vision branch first produces single-modal (i.e., RGB or thermal) detections. Denote the prediction scores as $S_{\scriptscriptstyle RGB}$, $S_{\scriptscriptstyle T}$ and bounding boxes as $B_{\scriptscriptstyle RGB}$, $B_{\scriptscriptstyle T}$ produced from each single-modal detector. The single-modal detections in the vision branch are processed through the language branch.  \\
\begin{figure*}[t]
\centering
\includegraphics[width=0.95\textwidth]{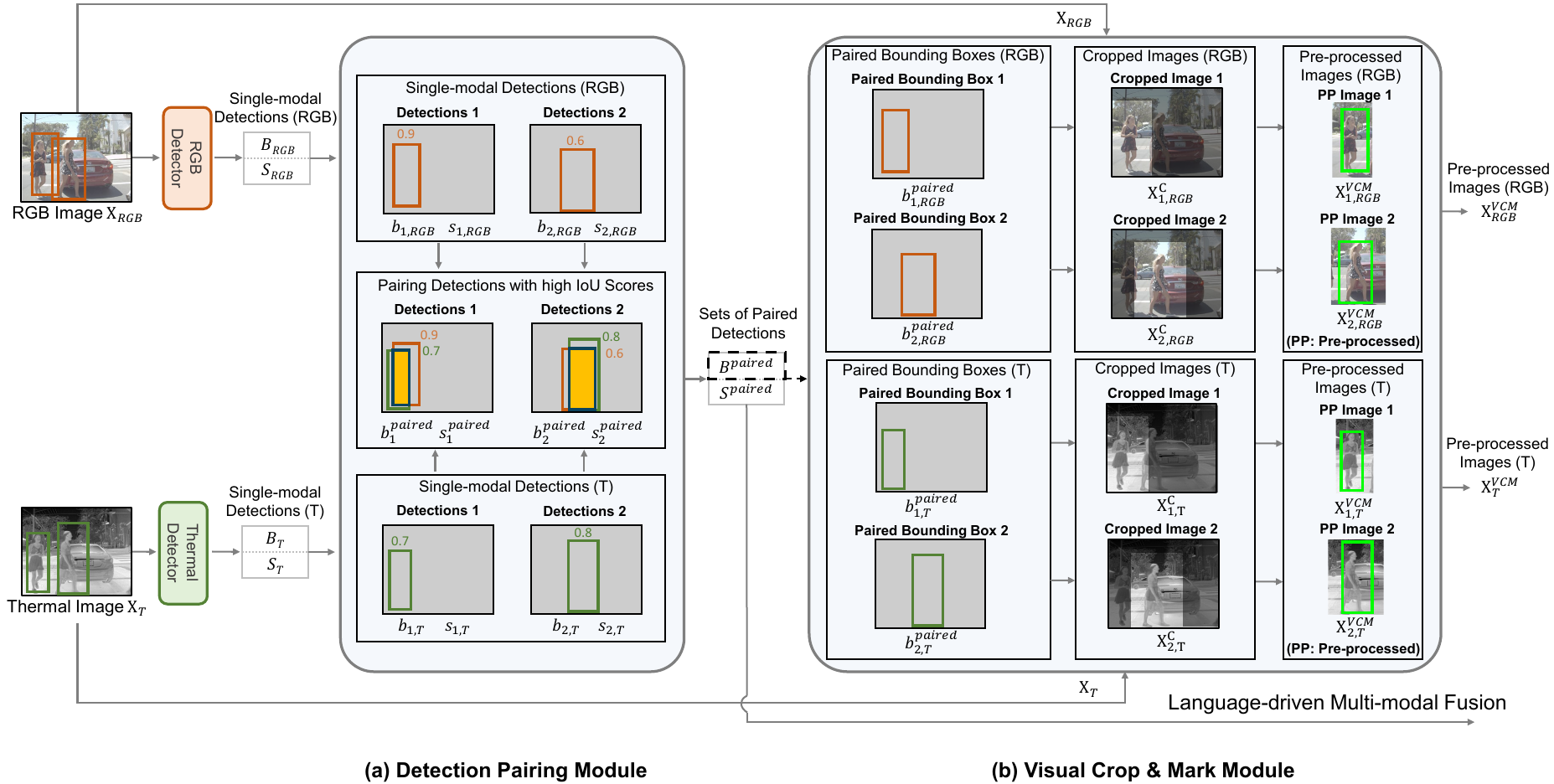}
\caption{Visualized details of the (a) Detection Pairing Module and the (b) Visual Crop \& Mark Module. (a) The Detection Pairing Module gets single-modal detections from the vision branch and then finds the detection pairs that belong to the same pedestrians, e.g., $b_{1}^{paired}$= ($b_{1,\scriptscriptstyle RGB}^{paired}$,$b_{1,\scriptscriptstyle T}^{paired}$) and $s_{1}^{paired}$= ($s_{1,\scriptscriptstyle RGB}^{paired}$,$s_{1,\scriptscriptstyle T}^{paired}$). Through the iteration of pedestrians, the module produces the sets of paired detections $B^{paired}$ and $S^{paired}$. (b) The Visual Crop \& Mark Module gets $B^{paired}$ as inputs, and output the pre-processed images $X_{\scriptscriptstyle RGB}^{\scriptscriptstyle VCM}$ and $X_{\scriptscriptstyle T}^{\scriptscriptstyle VCM}$.}
\label{fig:3}
\end{figure*}
\indent \textbf{Language Branch.} The language branch first generates text descriptions of pedestrians and performs detections based on them. Toward this goal, we leverage Multi-modal Large Language Models (MLLMs), which are capable of generating comprehensive descriptions in both RGB and thermal images. Based on these descriptions, we propose the Multispectral Chain-of-Thought (MSCoT) prompting, which prompts the LLM to conduct reasoning steps and produce detections (Section \ref{sec:3.3}). Detections from MSCoT prompting include single-modal (i.e., RGB and thermal) prediction scores  $S_{\scriptscriptstyle RGB}^{\scriptscriptstyle L}$, $S_{\scriptscriptstyle T}^{\scriptscriptstyle L}$, and fused prediction scores $S_{\scriptscriptstyle F}^{\scriptscriptstyle L}$. In Sections \ref{sec:3.2} and \ref{sec:3.3}, we describe the processes to produce these detections in detail.\\
\indent \textbf{Language-driven Multi-modal Fusion (LMF).} Vision-driven detection, language-driven detection, and vision-language fusion are performed to produce final detections, $S_{\scriptscriptstyle LMF}$ and  $B_{\scriptscriptstyle LMF}$. In Section \ref{sec:3.4}, we describe the fusion strategies in detail.
\subsection{Cross-modal Pedestrian Description Generation}
\label{sec:3.2}
The first step of the language branch is to generate text descriptions of pedestrians, both from the RGB and thermal images. We prompt a Multi-modal Large Language Model (MLLM) ChatGPT-4V~\cite{achiam2023gpt} to describe pedestrian regions in the image, which can be identified by their bounding box coordinates $B_{\scriptscriptstyle RGB}$, $B_{\scriptscriptstyle T}$ obtained from single-modal detectors. However, there are two challenges, which we introduce below with their solutions.\\
\indent \textbf{Detection Pairing.} The first challenge is due to the misalignment in multispectral data~\cite{hwang2015multispectral,kim2021uncertainty}. The same pedestrian often appears in different locations within the RGB and thermal image pairs, as their image sensors may have different Field-of-View or frame rates~\cite{kim2021uncertainty}. In such cases, we need to guide MLLMs to make descriptions based on different image regions in RGB and thermal images for the same pedestrians. Our strategy is to find the detection pairs from the single-modal detections (e.g., bounding boxes) that belong to the same pedestrians and provide this information to the MLLM. \\ 

\begin{algorithm}[t!]
\SetAlgoLined 
\caption{Detection Pairing (DPair)}
\label{alg:dpair}
\begin{algorithmic}[1]
\small
\STATE \textbf{Input}: Prediction scores $S_{\scriptscriptstyle RGB}$, $S_{\scriptscriptstyle T}$, Bounding boxes $B_{\scriptscriptstyle RGB}$, $B_{\scriptscriptstyle T}$.
\STATE \textbf{Require}: An IoU threshold $\tau$.
\STATE \textbf{Output}: Paired bounding boxes $B^{paired}$, paired prediction scores $S^{paired}$.\\ 
\STATE \textbf{procedure} DPair$(B_{\scriptscriptstyle RGB},B_{\scriptscriptstyle T},S_{\scriptscriptstyle RGB},S_{\scriptscriptstyle T};\tau)$
\STATE \indent \textbf{initialize} lists $B \gets B_{\scriptscriptstyle RGB} \cup B_{\scriptscriptstyle T}$, $S \gets S_{\scriptscriptstyle RGB} \cup S_{\scriptscriptstyle T}$, \indent $B^{paired} \gets \phi$, and $S^{paired} \gets \phi$.
    \STATE \indent Find the box $b_{max} \in B$ of the highest prediction score.
    \STATE \indent Find boxes $D\in B$ overlapping with $b_{max}$, of IoU ($>$$\tau$) \indent in the opposite modality. 
    \STATE \indent Find the box $b_{pair} \in D$ of the highest IoU value.
    \STATE \indent If $D$ is empty, then $b_{pair}$ $\gets$ $b_{max}$ and $s_{pair}$ $\gets$ $s_{max}$.
    \STATE \indent Append ($b_{max}, b_{pair}$) to $B^{paired}$.
    \STATE \indent Append ($s_{max}, s_{pair}$) to $S^{paired}$.
\STATE \indent Remove $b_{max}$ and $b_{pair}$ from $B$ and $s_{max}$ and $s_{pair}$ \indent from $S$
\STATE \indent Repeat this process until \(B\) and \(S\) are empty.
\STATE \indent $\textbf{return}$ $B^{paired}$, $S^{paired}$ 
\STATE $\textbf{end procedure}$
\end{algorithmic}
\end{algorithm}

\begin{algorithm}[t!]
\caption{Visual Crop \& Mark (VCM)}
\label{alg:vcm}
\begin{algorithmic}[1]
\small
\STATE \textbf{Input}: An image $X$, and bounding boxes $B$.
\STATE \textbf{Output}: Pre-processed images $X^{VCM}$.\\ 
\STATE \textbf{procedure} VCM$(X,B)$  
\STATE \indent \textbf{initialize} ${X^{\scriptscriptstyle VCM}} \gets \phi $.
\STATE \indent \textbf{for} all bounding boxes $b$ in $B$
\State \indent \indent Draw a green box at bounding box $b$ in the \indent \indent image $X$.
\State \indent \indent Crop the image $X$ around the bounding box $b$ \indent\indent such that the width and height of the cropped \indent\indent image $X_{c}$ are twice as the width and height of $b$.
\State \indent \indent Append the cropped image \(X_c\) to \(X^{\scriptscriptstyle VCM}\).
\STATE \indent \textbf{end for}
\STATE \indent $\textbf{return}$ $X^{\scriptscriptstyle VCM}$ 
\STATE $\textbf{end procedure}$ 
\end{algorithmic}
\end{algorithm}
\vspace{-0.2cm}
\indent Although bounding boxes representing the same pedestrian can have different coordinates in different modalities, still they will contain highly overlapping areas in the images, and those boxes will have large values of IoUs (Intersection-of-Unions). Specifically, for the $i$-th bounding box $b_{\scriptscriptstyle i,\scriptscriptstyle m}$ in bounding boxes $B_{\scriptscriptstyle m}$, obtained from the single-modal detector in the $m$ modality ($m$ is either RGB or thermal: T), we aim to find the bounding box from $B_{\scriptscriptstyle m^{c}}$, in the opposite modality $m^{c}$ that corresponds to the same pedestrian. To this end, we compute the IoU value between $b_{\scriptscriptstyle i,\scriptscriptstyle m}$ across all bounding boxes in $B_{\scriptscriptstyle m^{c}}$, and find the box that has the highest IoU value, e.g., $b_{\scriptscriptstyle j,\scriptscriptstyle m^{c}}$, with $b_{\scriptscriptstyle i,\scriptscriptstyle m}$. Here, we only consider the boxes with IoU values above the IoU threshold $\tau$. With indices $i$ and $j$, we can find a pair of the prediction scores $s_{\scriptscriptstyle i,\scriptscriptstyle m}$ and $s_{\scriptscriptstyle j,\scriptscriptstyle m^{c}}$ that belong to the same pedestrian. We call such tuples ($b_{\scriptscriptstyle i,\scriptscriptstyle m}$,$b_{\scriptscriptstyle j,\scriptscriptstyle m^{c}}$) and ($s_{\scriptscriptstyle i,\scriptscriptstyle m}$,$s_{\scriptscriptstyle j,\scriptscriptstyle m^{c}}$) as the ``paired detection". When all IoU values are under the IoU threshold $\tau$, this means that there are no boxes in $B_{\scriptscriptstyle m^{c}}$ that correspond to the same pedestrian and $b_{\scriptscriptstyle i,\scriptscriptstyle m}$ can not find a pair. Such cases can occur at nighttime or in thermal-obscured scenarios where the pedestrian is detected by only one modality. In this case, we override the value of $b_{\scriptscriptstyle i,\scriptscriptstyle m}$ and $s_{\scriptscriptstyle i,\scriptscriptstyle m}$ to the opposite modality, making pairs of ($b_{\scriptscriptstyle i,\scriptscriptstyle m}$,$b_{\scriptscriptstyle i,\scriptscriptstyle m}$) and ($s_{\scriptscriptstyle i,\scriptscriptstyle m}$,$s_{\scriptscriptstyle i,\scriptscriptstyle m}$). Iterating this process for over all bounding boxes in $B_{\scriptscriptstyle m}$ and $B_{\scriptscriptstyle m^{c}}$ (i.e., $B_{\scriptscriptstyle RGB}$ and $B_{\scriptscriptstyle T}$) allows us to obtain a set of paired detections $B^{paired}$ and $S^{paired}$. Denote the Detection Pairing Module as a function $DPair(\cdot)$ with respect to inputs $B_{\scriptscriptstyle RGB}$, $B_{\scriptscriptstyle T}$, $S_{\scriptscriptstyle RGB}$, and $S_{\scriptscriptstyle T}$, which indicate bounding boxes and prediction scores obtained from RGB and thermal single-modal detectors, respectively. The inputs and outputs of $DPair(\cdot)$ can be represented as:
\begin{equation}{
S^{paired}, B^{paired}=DPair(B_{\scriptscriptstyle RGB},B_{\scriptscriptstyle T},S_{\scriptscriptstyle RGB},S_{\scriptscriptstyle T};\ \tau),
\label{eq:4}
}\end{equation}
where $\tau$ indicates an IoU threshold. We set this value to 0.5, the standard value used in the object detection task. $B^{paired}$ is used to guide MLLMs the location of when making descriptions in RGB and thermal images. $S^{paired}$ is used when fusing prediction scores from vision and language branches in Section \ref{sec:3.4}. The complete algorithm of the Detection Pairing Module is described in Algorithm 1.\\
\indent \textbf{Visual Crop \& Mark (VCM).}
Another challenge when using MLLMs for making descriptions is the low accuracy of MLLMs on small-scale objects, i.e., objects occupying a small area in the image. To solve this issue, we refer to a recent method~\cite{zhang2023visual} that improves the performance of MLLMs on small visual details by pre-processing the input image with image cropping. Similarly, we zoom in on the target pedestrians and crop the image around it, such that visual information irrelevant to the pedestrian is removed. We find such a pre-processing strategy is shown to significantly improve the accuracy of MLLMs in describing small-scale pedestrians, near the level comparable to recognizing large-scale pedestrians.\\ 
\indent The detailed algorithm for VCM is described in Algorithm 2. First, an image $X$ and bounding boxes $B$ are given as input. Note that $B$ represents the locations of pedestrians in image $X$. For each bounding box $b$, the image $X$ is cropped around $b$ so that the width and height of the cropped image $X{c}$ are twice as $b$.\\
\begin{algorithm}[t!]
\SetAlgoLined 
\caption{Cross-aligned Pedestrian Description Generation (CPDG)}
\label{cpdg}
\begin{algorithmic}[1]
\small
\STATE \textbf{Input}: An RGB image $X_{\scriptscriptstyle RGB}$, an thermal image $X_{\scriptscriptstyle T}$, prediction scores $S_{\scriptscriptstyle RGB}$, $S_{\scriptscriptstyle T}$ and bounding boxes $B_{\scriptscriptstyle RGB}$, $B_{\scriptscriptstyle T}$. 
\STATE \textbf{Require}: Input text prompts $p_{\scriptscriptstyle RGB}$, $p_{\scriptscriptstyle T}$, and IoU threshold $\tau$. Detection pairing module DPair($\cdot$), Visual Crop \& Mark module VCM($\cdot$). A pre-trained multimodal large language model MLLM($\cdot$).
\STATE \textbf{Output}: A set of RGB descriptions $D_{RGB}$, a set of thermal descriptions $D_{T}$, and paired detections $S^{paired}$, $B^{paired}$.\\ 
\STATE \textbf{procedure} CPDG$(X_{\scriptscriptstyle RGB},X_{\scriptscriptstyle T},S_{\scriptscriptstyle RGB},S_{\scriptscriptstyle T},B_{\scriptscriptstyle RGB},B_{\scriptscriptstyle T},$ \
$p_{\scriptscriptstyle RGB},p_{\scriptscriptstyle T},\tau)$
\State \indent $S^{paired}$, $B^{paired}$ $\gets$ DPair$(B_{\scriptscriptstyle RGB},B_{\scriptscriptstyle T},S_{\scriptscriptstyle RGB},S_{\scriptscriptstyle T};\tau)$.
\STATE \indent $X_{\scriptscriptstyle RGB}^{\scriptscriptstyle VCM}$ $\gets$ VCM($X_{\scriptscriptstyle RGB}$, $B^{paired}$). \\ \indent \indent $X_{\scriptscriptstyle T}^{\scriptscriptstyle VCM}$ $\gets$ VCM($X_{\scriptscriptstyle T}$, $B^{paired}$).
\STATE \indent \textbf{initialize} list $D_{RGB} \gets \phi, D_{T} \gets \phi$.
\STATE \indent \textbf{for} all images $x^{\scriptscriptstyle VCM}_{RGB}$ in $X^{\scriptscriptstyle VCM}_{RGB}$
\State \indent \indent $d_{RGB}\gets$MLLM$(x^{\scriptscriptstyle VCM}_{RGB};p_{RGB})$.
\State \indent \indent Append the text description \(d_{RGB}\) to \(D_{RGB}\).
\STATE \indent \textbf{end for}
\STATE \indent \textbf{for} all images $x^{\scriptscriptstyle VCM}_{T}$ in $X^{\scriptscriptstyle VCM}_{T}$
\State \indent \indent $d_{T}\gets$MLLM$(x^{\scriptscriptstyle VCM}_{T};p_{T})$.
\State \indent \indent Append the text description \(d_{T}\) to \(D_{T}\).
\STATE \indent \textbf{end for}
\STATE \indent $\textbf{return}$ $D_{RGB}$, $D_{T}$, $S^{paired}$, $B^{paired}$.
\STATE $\textbf{end procedure}$ 
\end{algorithmic}
\end{algorithm}
\indent In addition to cropping, we visually guide the MLLM to describe a specific pedestrian by drawing a 1-pixel width ``green" color (with RGB values (0,255,0)) box around a target pedestrian. We use the ``green" color to draw the boxes because it is most distinctly in hue from human skin color, thus effective for distinguishing humans from the background. This process is iterated for all bounding boxes $b \in B$ for both a RGB image $X_{\scriptscriptstyle RGB}$ and a thermal image $X_{\scriptscriptstyle T}$. Denote this the Visual Crop \& Mark process as $VCM(\cdot)$, then the pre-processed RGB and thermal images $X_{\scriptscriptstyle RGB}^{\scriptscriptstyle VCM}$ and $X_{\scriptscriptstyle T}^{\scriptscriptstyle VCM}$ can be written as: 
\begin{equation}{
X_{\scriptscriptstyle RGB}^{\scriptscriptstyle VCM} \gets VCM(X_{\scriptscriptstyle RGB}, B^{paired}),
\label{eq:5}
}\end{equation}
\begin{equation}{
X_{\scriptscriptstyle T}^{\scriptscriptstyle VCM} \gets VCM(X_{\scriptscriptstyle T}, B^{paired}),
\label{eq:11}
}\end{equation}
where $X_{\scriptscriptstyle RGB}$ and $X_{\scriptscriptstyle T}$ are RGB and thermal input images, and $B^{paired}$ is the set of the paired bounding boxes. When pre-processing images in $X_{\scriptscriptstyle RGB}$ and $X_{\scriptscriptstyle T}$, bounding boxes of the corresponding modality in $B^{paired}$ are used only. \\
\indent \textbf{Generating Descriptions with MLLMs.} For each pre-processed image $x_{\scriptscriptstyle RGB}^{\scriptscriptstyle VCM}\in X_{\scriptscriptstyle RGB}^{\scriptscriptstyle VCM}$ and $x_{\scriptscriptstyle T}^{\scriptscriptstyle VCM}\in X_{\scriptscriptstyle T}^{\scriptscriptstyle VCM}$, we prompt the MLLM as the following. For the RGB image, \textit{``In this RGB image, what is in the green box?"} and for the thermal image, \textit{``In this thermal image, what is in the green box?"}. Denote these prompts as $p_{\scriptscriptstyle RGB}$, and $p_{\scriptscriptstyle T}$. Then, we can generate their corresponding text descriptions $d_{\scriptscriptstyle RGB}$ and $d_{\scriptscriptstyle T}$ such as:
\begin{equation}{
d_{\scriptscriptstyle RGB} \gets MLLM(x_{\scriptscriptstyle RGB}^{\scriptscriptstyle VCM}; \ p_{\scriptscriptstyle RGB}),
\label{eq:6}
}\end{equation}
\begin{equation}{
d_{\scriptscriptstyle T} \gets MLLM(x_{\scriptscriptstyle T}^{\scriptscriptstyle VCM}; \ p_{\scriptscriptstyle T}),
\label{eq:16}
}\end{equation}
where $MLLM(\cdot)$ denotes the process of generating text descriptions from an MLLM, given the pre-processed image as input. Text descriptions are generated for all $x_{\scriptscriptstyle RGB}^{\scriptscriptstyle VCM}\in X_{\scriptscriptstyle RGB}^{\scriptscriptstyle VCM}$ and $x_{\scriptscriptstyle T}^{\scriptscriptstyle VCM}\in X_{\scriptscriptstyle T}^{\scriptscriptstyle VCM}$, producing sets of descriptions $ D_{\scriptscriptstyle RGB}$ and $D_{\scriptscriptstyle T}$. The detailed algorithm for Cross-modal Pedestrian Description Generation (CPDG) is described in Algorithm 3.
\begin{figure*}[t]
\centering
\includegraphics[width=0.8\textwidth]{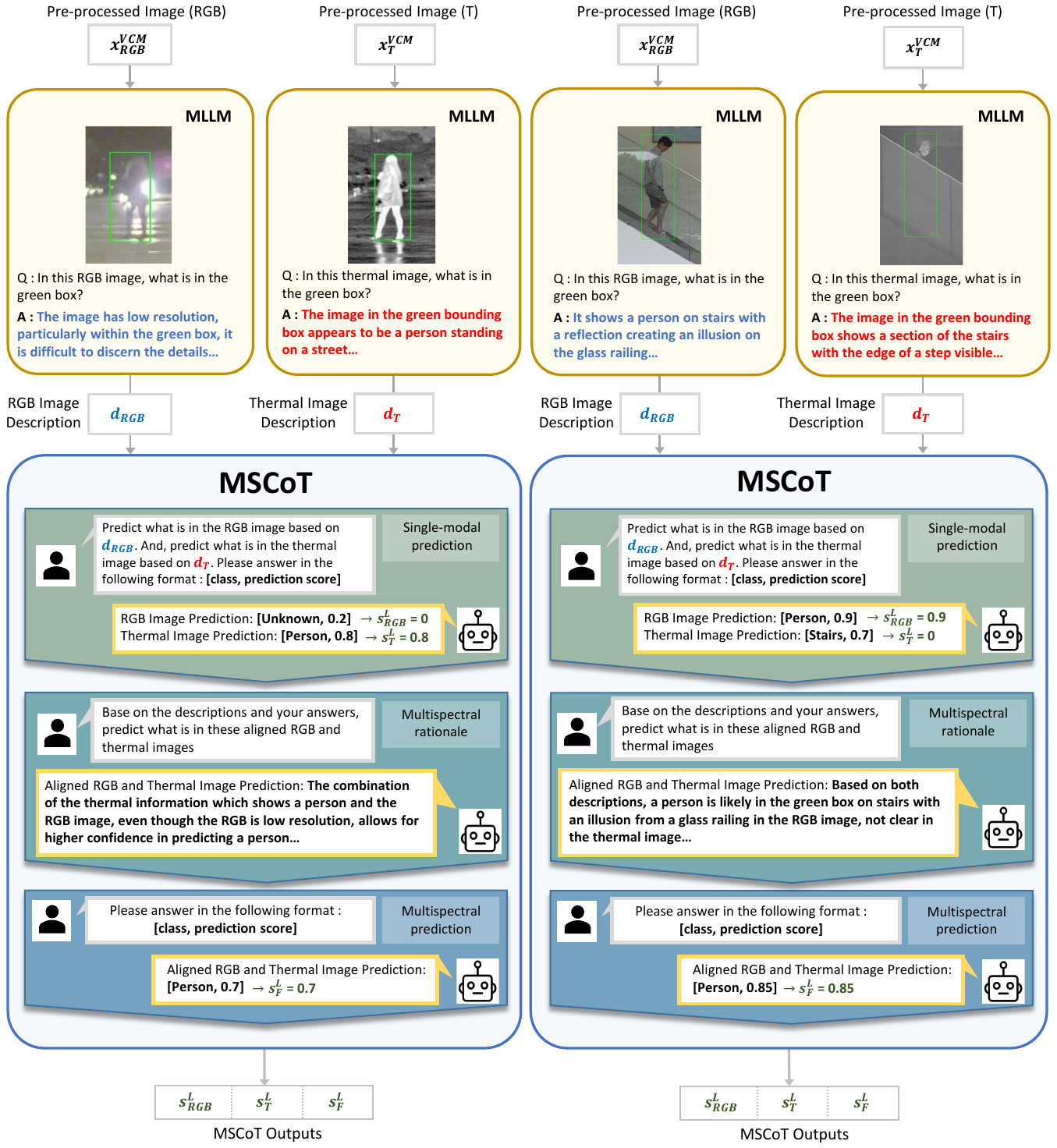}
\caption{Visualized process of our proposed Multispectral Chain-of-Thought (MSCoT) prompting.} 
\label{fig:1}
\end{figure*}
\vspace{-0.3cm}
\subsection{Multispectral Chain-of-Thought Prompting}
\label{sec:3.3}
We introduce Multispectral Chain-of-Thought (MSCoT) prompting, which facilitates LLMs to build step-by-step cross-modal rationales and produce detection results. MSCoT prompting first comprehends intra-modality information and produces single-modal prediction scores from each RGB and thermal description. Then, MSCoT prompting comprehends inter-modality reasoning and outputs fused detection scores and bounding boxes. This multi-step reasoning process addresses LLMs' overconfidence~\cite{xiong2023can} and produce calibrated confidence scores. \\
\indent To this end, we refer to the Chain-of-Thought (CoT) prompting~\cite{wei2022chain} technique, which significantly improves LLM's reasoning ability in complex NLP tasks. CoTs add a series of intermediate reasoning steps into LLMs and generate rationales when predicting answers. Motivated by the late-fusion process in multispectral pedestrian detection, we propose a Multispectral Chain-of-Thought (MSCoT) prompting chain. \\
\indent Given the text descriptions $d_{\scriptscriptstyle RGB}\in D_{\scriptscriptstyle RGB}$ and $d_{\scriptscriptstyle T}\in D_{\scriptscriptstyle T}$ of a RGB image and a thermal image, MSCoT prompting first prompts the LLM to produce single-modal prediction scores $s_{\scriptscriptstyle RGB}^{\scriptscriptstyle L}$, and $s_{\scriptscriptstyle T}^{\scriptscriptstyle L}$ and then predict a fused prediction score $s_{\scriptscriptstyle F}^{\scriptscriptstyle L}$ based on rationale chains. Given the descriptions $d_{\scriptscriptstyle RGB}$, and $d_{\scriptscriptstyle T}$ of RGB and thermal images as the context, we design the following prompt chains. \\
\\
\indent \textbf{Generating Single-modal Prediction.}\\
Given the RGB image description $d_{\scriptscriptstyle RGB}$ and the thermal image description $d_{\scriptscriptstyle T}$, the LLM is prompted with context $[d_{\scriptscriptstyle RGB} \ d_{\scriptscriptstyle T}]$ and text prompt $p_{single}$ as the following (line 8 in Algorithm.4).\\ \\
Context $[d_{\scriptscriptstyle RGB} \ d_{\scriptscriptstyle T}]$: RGB image description $d_{\scriptscriptstyle RGB}$, thermal image description $d_{\scriptscriptstyle T}$.\\
Prompt $p_{single}$: First, predict what is in the RGB image based on $d_{\scriptscriptstyle RGB}$. And predict what is in the thermal image based on $d_{\scriptscriptstyle T}$. Please answer in the format : [class, prediction score]\\
\\
From the above prompt, the LLM outputs \{[$c_{RGB}^{L}$, $s_{RGB}^{L}$],
[$c_{T}^{L}$, $s_{T}^{L}$]\}, including single-modal prediction scores $s_{\scriptscriptstyle RGB}^{\scriptscriptstyle L}$, $s_{\scriptscriptstyle T}^{\scriptscriptstyle L}$ and class labels $c_{\scriptscriptstyle RGB}^{\scriptscriptstyle L}$, $c_{\scriptscriptstyle T}^{\scriptscriptstyle L}$ (line 8 in Algorithm. 4). $s_{\scriptscriptstyle RGB}^{\scriptscriptstyle L}$, $s_{\scriptscriptstyle T}^{\scriptscriptstyle L}$ are assigned zero if $c_{\scriptscriptstyle RGB}^{\scriptscriptstyle L}$, $c_{\scriptscriptstyle T}^{\scriptscriptstyle L}$ are not ``person", respectively (line 9-10 in Algorithm. 4).
\\
\\
\indent \textbf{Generating Multispectral Rationales and Prediction.}\\
Additionally providing $s_{\scriptscriptstyle RGB}^{\scriptscriptstyle L}$, and $s_{\scriptscriptstyle T}^{\scriptscriptstyle L}$ as the context, the LLM is prompted to generate multispectral rationales and fused predictions. Given the RGB image description $d_{RGB}$, the thermal image description $d_{T}$, and answers $s_{\scriptscriptstyle RGB}^{\scriptscriptstyle L}$, $s_{\scriptscriptstyle T}^{\scriptscriptstyle L}$ of the single-modal predictions, the LLM is prompted with the context $[d_{RGB} \ d_{T} \ s_{\scriptscriptstyle RGB}^{\scriptscriptstyle L} \ s_{\scriptscriptstyle T}^{\scriptscriptstyle L}]$ and text prompt $p_{multi}$ as the following (line 12 in Algorithm.4).\\ \\
Context $[d_{RGB} \ d_{T} \ s_{\scriptscriptstyle RGB}^{\scriptscriptstyle L} \ s_{\scriptscriptstyle T}^{\scriptscriptstyle L}]$: RGB image description: $d_{\scriptscriptstyle RGB}$, thermal image description: $d_{\scriptscriptstyle T}$, answers: $s_{\scriptscriptstyle RGB}^{\scriptscriptstyle L}$, and $s_{\scriptscriptstyle T}^{\scriptscriptstyle L}$.\\
Prompt $p_{multi}$: Based on the descriptions and your answers, predict what is in these aligned RGB and thermal images. Please answer in the format: [class, prediction score]\\
\\
From the above prompt, the LLM outputs [$c_{F}^{L}$, $s_{F}^{L}$], including fused prediction score $s_{\scriptscriptstyle F}^{\scriptscriptstyle L}$ and class label $c_{\scriptscriptstyle F}^{\scriptscriptstyle L}$ (line 12 in Algorithm. 4). $s_{\scriptscriptstyle F}^{\scriptscriptstyle L}$, is assigned zero if $c_{\scriptscriptstyle F}^{\scriptscriptstyle L}$ is not ``person". (line 13 in Algorithm. 4). $s_{\scriptscriptstyle RGB}^{\scriptscriptstyle L}$, $s_{\scriptscriptstyle T}^{\scriptscriptstyle L}$, and $s_{\scriptscriptstyle F}^{\scriptscriptstyle L}$, the outputs of MSCoT Prompting with respect to the inputs $d_{\scriptscriptstyle RGB}$, $d_{\scriptscriptstyle T}$ can expressed as:
\begin{equation}{
s_{\scriptscriptstyle RGB}^{\scriptscriptstyle L}, s_{\scriptscriptstyle T}^{\scriptscriptstyle L},s_{\scriptscriptstyle F}^{\scriptscriptstyle L}=MSCoT(d_{\scriptscriptstyle RGB}, \ d_{\scriptscriptstyle T}).
}
\label{eq:7}
\end{equation}
\indent Iterating this process over all $d_{\scriptscriptstyle RGB}\in D_{\scriptscriptstyle RGB}$ and $d_{\scriptscriptstyle T}\in D_{\scriptscriptstyle T}$ obtains the prediction scores: $S_{\scriptscriptstyle RGB}^{\scriptscriptstyle L}$, $S_{\scriptscriptstyle T}^{\scriptscriptstyle L}$, and $S_{\scriptscriptstyle F}^{\scriptscriptstyle L}$. Our implementation for MSCoT prompting is achieved by fine-tuning ChatGPT-3.5~\cite{chatgpt} with about 50 training samples. The implementation details are described below.
\begin{algorithm}[t!]
\SetAlgoLined 
\caption{Multispectral Chain-of-Thought (MSCoT) Prompting} 
\label{mscot}
\begin{algorithmic}[1]
\small
\STATE \textbf{Input}: RGB image description $d_{RGB}$, and thermal image description $d_{T}$. 
\STATE \textbf{Require}: A pre-trained large language model LLM($\cdot$), \\
text prompts $p_{single}$ and $p_{multi}$.
\STATE \textbf{Output}: RGB single-modal prediction score 
$s_{RGB}^{L}$, thermal single-modal prediction score $s_{T}^{L}$, fused prediction score $s_{F}^{L}$.\\ 
\STATE \textbf{procedure} MSCoT($d_{RGB}$, $d_{T}$)
\State \indent $\#$ Generate Single-Modal Predictions
\State \indent \{[$c_{RGB}^{L}$, $s_{RGB}^{L}$],
[$c_{T}^{L}$, $s_{T}^{L}$]\}$\gets LLM([d_{RGB} \ d_{T}];p_{single})$.
\State \indent $s_{RGB}^{L}\gets$ 0 if $c_{RGB}^{L}\mathrel{\mathtt{!=}}$``person"
\State \indent $s_{T}^{L}\gets$ 0 if $c_{T}^{L}\mathrel{\mathtt{!=}}$``person"
\State \indent $\#$ Generate Multispectral Rationales and Predictions
\State \indent [$c_{F}^{L}$, $s_{F}^{L}]\gets$LLM$([d_{RGB}\ d_{T}\ s_{RGB}\ s_{T}];p_{multi})$.
\State \indent $s_{F}^{L}\gets$ 0 if $c_{F}^{L}\mathrel{\mathtt{!=}}$``person"
\STATE \indent $\textbf{return}$ $s_{RGB}^{L}$, $s_{T}^{L}$, $s_{F}^{L}$.
\STATE $\textbf{end procedure}$
\end{algorithmic}
\end{algorithm}
\indent \textbf{Finetuning the LLM for MSCoT.} We use the ChatGPT-3.5 (GPT-3.5) API~\cite{chatgpt}, provided by the OpenAI developer platform. Using the fine-tuning module provided on the GPT official website~\cite{gptfinetuning}, we trained GPT-3.5 API with our training data. The official document for GPT-3.5~\cite{gptfinetuning} recommends training 50 samples for fine-tuning the model, and we utilized a total of 50 training samples for each FLIR and CVC-14 model. The training data contains textual descriptions of RGB and thermal images, rationale, and prediction scores for training the Large Language Model. We prompt the GPT-3.5~\cite{chatgpt} consisting of rationale and prediction scores. The following answers are obtained by prompting the GPT-3.5 independently with these prompts, and we merge the answers to make the training data. Then with these data, we fine-tune the model answer based on a chain of prompts. The training data are manually selected with high-quality answers, containing reasonable rationales and confidence scores. 50 training data from the FLIR training images are used for fine-tuning the MSCoT of the FLIR model. Also, we made 50 training data using the CVC-14 training images for MSCoT of the CVC-14 model. Note that images from ROTX-MP are not used for making training data for MSCoT because it violates the evaluation setting for ROTX-MP.\\
\subsection{Language-driven Multi-modal Fusion (LMF)}
\label{sec:3.4}
We propose to ensemble the detections of the vision branch and the language branch at the last stage of the proposed network (Fig. \ref{overall_arch}). The Language-driven Multi-modal Fusion (LMF) consists of three implementation mechanisms: 1) Vision-driven Detection (Eq.10), 2) Language-driven Detection (Eq.11), and 3) Vision-Language Fusion (Eq.12). Vision-driven Detection (Eq.10) and Language-driven Detection (Eq.11) calculate detection scores and bounding boxes separately for each vision and language branch. Vision-Language Fusion (Eq.12) then fuses detection results from the vision branch and the language branch.\\
\indent \textbf{Vision-driven Detection.} For the vision branch, denote $S_{\scriptscriptstyle RGB}^{\scriptscriptstyle V}$ as the prediction scores from $S^{paired}$, and $B_{\scriptscriptstyle RGB}^{\scriptscriptstyle V}$ as the bounding boxes from $B^{paired}$ that corresponds to the RGB modality, similar for $S_{\scriptscriptstyle T}^{\scriptscriptstyle V}$ and $B_{\scriptscriptstyle T}^{\scriptscriptstyle V}$. Then, the vision-driven detections $S_{\scriptscriptstyle F}^{\scriptscriptstyle V}$ (predictions scores) and $B_{\scriptscriptstyle F}^{\scriptscriptstyle V}$ (bounding boxes) can be produced as:
\begin{equation}{
S_{\scriptscriptstyle F}^{\scriptscriptstyle V}=max(S_{\scriptscriptstyle RGB}^{\scriptscriptstyle V}\ , \ S_{\scriptscriptstyle T}^{\scriptscriptstyle V}), \ B_{\scriptscriptstyle F}^{\scriptscriptstyle V}= \frac{B_{\scriptscriptstyle RGB}^{\scriptscriptstyle V}\odot S_{\scriptscriptstyle RGB}^{\scriptscriptstyle V} + B_{\scriptscriptstyle T}^{\scriptscriptstyle V}\odot S_{\scriptscriptstyle T}^{\scriptscriptstyle V}}{S_{\scriptscriptstyle RGB}^{\scriptscriptstyle V} + S_{\scriptscriptstyle T}^{\scriptscriptstyle V}},
\label{eq:8}
}\end{equation}
where $\odot$ denotes the Hadamard product (element-wise multiplication).
NMS (Eq. \ref{eq:2}) and weighted-averaging (Eq. \ref{eq:3}) are applied for score-fusion and box-fusion. \\
\indent \textbf{Language-driven Detection.} Next, for the language branch, fused bounding boxes $B_{\scriptscriptstyle F}^{\scriptscriptstyle L}$ can be produced as:
\begin{equation}{
B_{\scriptscriptstyle F}^{\scriptscriptstyle L}=\frac{B_{\scriptscriptstyle RGB}^{\scriptscriptstyle V}\odot S_{\scriptscriptstyle RGB}^{\scriptscriptstyle L} + B_{\scriptscriptstyle T}^{\scriptscriptstyle V}\odot S_{\scriptscriptstyle T}^{\scriptscriptstyle L}}{S_{\scriptscriptstyle RGB}^{\scriptscriptstyle L} + S_{\scriptscriptstyle T}^{\scriptscriptstyle L}},
\label{eq:9}
}\end{equation}
given the bounding boxes $B_{\scriptscriptstyle RGB}^{\scriptscriptstyle V}$, $B_{\scriptscriptstyle T}^{\scriptscriptstyle V}$ obtained single-modal detectors. 
Weighted-averaging (Eq. \ref{eq:3}) is applied for box-fusion. The fused prediction scores $S_{\scriptscriptstyle F}^{\scriptscriptstyle L}$ are produced by the MSCoT prompting (Section \ref{sec:3.3}). \\
\indent \textbf{Vision-Language Fusion.} The LMF fuse detection results from the vision branch and the language branch by Vision-Language Fusion (Eq.12). Vision-Language Fusion is particularly designed to handle potential information asymmetries by weighting the detection scores $S_{\scriptscriptstyle F}^{\scriptscriptstyle V}$, $S_{\scriptscriptstyle F}^{\scriptscriptstyle L}$ and bounding box coordinates $B_{\scriptscriptstyle F}^{\scriptscriptstyle V}$, $B_{\scriptscriptstyle F}^{\scriptscriptstyle L}$ obtained from both branches. The final detections are produced as:
\begin{equation}{
S_{\scriptscriptstyle LMF}=avg(S_{\scriptscriptstyle F}^{\scriptscriptstyle V} , S_{\scriptscriptstyle F}^{\scriptscriptstyle L}), \ B_{\scriptscriptstyle LMF}=\frac{S_{\scriptscriptstyle F}^{\scriptscriptstyle V} \odot B_{\scriptscriptstyle F}^{\scriptscriptstyle V} + S_{\scriptscriptstyle F}^{\scriptscriptstyle L}\odot B_{\scriptscriptstyle F}^{\scriptscriptstyle L}}{S_{\scriptscriptstyle F}^{\scriptscriptstyle V} + S_{\scriptscriptstyle F}^{\scriptscriptstyle L}},
\label{eq:10}
}\end{equation}
where $S_{\scriptscriptstyle LMF}$ and $B_{\scriptscriptstyle LMF}$ denote final prediction scores and bounding boxes. Averaging (Eq. \ref{eq:2}) and weighted-averaging (Eq. \ref{eq:3}) are applied for score-fusion and box-fusion.
\\
\indent \textbf{Design Choice.} For score-fusion, we adopted the NMS method (Eq. \ref{eq:2}) in the vision-driven detection. NMS compares the prediction scores estimated in each modality and votes for the highest one, removing the lower score. We apply such a strategy because higher confidence scores generally occur when there is more useful information for detecting the pedestrian. From this strategy, we can obtain higher prediction scores for true-positive detections in most cases. However, it can also increase the prediction scores for false positives. As MSCoT prompting can comprehend very low prediction scores for false positives, we adopt the averaging strategy (Eq. \ref{eq:1}) in the vision-language fusion to refine over-confident false positives from the vision-driven detections. For box-fusion, we adopted the weighting average (Eq. \ref{eq:3}) method in all processes, implying that more confident detections should have a higher weight when fusing. In Section \ref{sec:6.2}, we conduct an ablation study on these design choices. 
\section{Experimental Setup}
\subsection{Dataset and Evaluation Metric}
\sloppy The experiments are conducted on multispectral pedestrian datasets: Teledyne FLIR Free ADAS Thermal Dataset v2.0.0~\cite{c:25}, CVC-14~\cite{gonzalez2016pedestrian}, ROTX-MP~\cite{kim2024causal}. 
The FLIR~\cite{c:25} dataset consists of RGB and thermal image pairs with an image resolution of $640 \times 512$. For a fair comparison with previous studies~\cite{zhang2020multispectral, qingyun2022crossmodality,kim2023multispectral,kim2024causal}, we use the aligned version of FLIR, proposed by~\cite{zhang2020multispectral}. This version filters out misaligned images, containing well-aligned 4,129 training and 1,013 test RGB and thermal image pairs. For convenience, we call this version of the dataset as FLIR. We evaluate both day and night images and report the performance on the entire test set (`All'). In contrast to FLIR, the CVC-14~\cite{gonzalez2016pedestrian} dataset often contains heavily misaligned pairs of RGB and thermal images. The train and test set each contain 3,618 and 1,433 grayscale and thermal images with a $640 \times 471$ resolution. We evaluate daytime (`Day'), nighttime (`Night'), and the total (`All') test images separately, following previous works~\cite{kim2021uncertainty, kim2024causal}.\\
Lastly, we evaluate our method on the ROTX-MP~\cite{kim2024causal} dataset, which consists of 1,000 test images of mainly thermal-obscured pedestrians. We use the models trained from FLIR and CVC-14 to test on ROTX-MP, as this dataset aims to evaluate multispectral pedestrian detectors when there is a substantial distribution change between the train and test splits. For the evaluation metric, we use the Average Precision (AP $\uparrow$). These experimental settings are the same as the original paper~\cite{kim2024causal}. 
\subsection{Implementation Detail}
\label{sec:4.2}
\indent \textbf{Single-modal Detectors.} For the single-modal detectors, we use the Co-DETR~\cite{zong2023detrs} model and train RGB images and thermal images separately. Co-DETR is based on the DETR (DEtection TRansformer)~\cite{carion2020endtoend} architecture, enhanced by collaborative learning with multiple parallel auxiliary heads integrated into the output of the transformer encoder. We train the single-modal detectors with RGB and thermal images from FLIR~\cite{c:25} and CVC-14~\cite{gonzalez2016pedestrian} training data, respectively. For optimizing Co-DETR, we use the same setting in the original paper~\cite{zong2023detrs}, using AdamW~\cite{loshchilov2017decoupled} optimizer with an initial learning rate of 1e-4 and weight decay of 1e-4. All models are trained on an eight NVIDIA A6000 GPU for 16 epochs with a batch size of 16. \\
\indent \textbf{Large Language Models (LLMs).} We use two types of Large Language Models. First is the Multimodal Large Language Model (MLLM), which we use to generate text descriptions for RGB and thermal pedestrian images (Section \ref{sec:3.2}). For the MLLM, we use the pre-trained model of the ChatGPT-vision (GPT-4V)~\cite{achiam2023gpt} API, provided by the OpenAI developer platform.  \\
\indent The second is the chatbot version, ChatGPT-3.5 (GPT-3.5) API~\cite{chatgpt}, also provided by the OpenAI developer platform. We fine-tuned the GPT-3.5 model for the Multispectral Chain-of-Thought (MSCoT) prompting (Section \ref{sec:3.3}) we proposed. Using the fine-tuning module provided on the GPT official website~\cite{gptfinetuning}, we trained GPT-3.5 API with our high-quality selection pairs consisting of RGB, thermal descriptions, rationale, and prediction scores. The official document for GPT-3.5~\cite{gptfinetuning} recommends training 50 samples for fine-tuning the model, and we utilized a total of 50 training samples.
\vspace{-0.4cm}
\subsection{Comparison Model}
We compare our method with seven multispectral pedestrian detection models: 1) Halfway Fusion~\cite{liu2016multispectral}, 2) Cross-modality Fusion Transformer (CFT)~\cite{qingyun2022crossmodality}, 3) Kim et al.~\cite{kim2021uncertainty}, 4) ProbEn~\cite{chen2022multimodal}, 5) Causal Mode Multiplexer (CMM)~\cite{kim2024causal}, 6) ICAFusion~\cite{shen2024icafusion}, and 7) MS-DETR~\cite{xing2024ms}. We used the same experimental settings as their original papers. \textbf{Halfway Fusion}: we use the Faster-RCNN~\cite{girshick2015fast} as the base model and train the model for 3 epochs. The learning rate is initialized at 0.008 during the initial 2 epochs with SGD~\cite{amari1993backpropagation} and subsequently applied a 0.1 learning rate decay for the final epoch.
\textbf{CFT}: The initial learning rate is 0.01 with a momentum of 0.937, and weight decay 0.0005. Batch size is 32 and the model is trained for 200 epochs and the YOLO-v5 weight trained on the COCO dataset is used for weight initialization. We used the official code. \textbf{Kim et al}: The learning rate is initialized at 0.006 during the initial 2 epochs and a 0.1 learning rate decay was applied for the final epoch. We used the official code. \textbf{ProbEn}: We finetune two single-modal detectors based on Faster R-CNN~\cite{girshick2015fast} pretrained on COCO~\cite{lin2014microsoft}. The Detectron 2~\cite{wu2019detectron2} library is used. We used SGD~\cite{amari1993backpropagation} optimizer with learning rate 5e-3. We adopt the `ProbEn' score fusion and `v-avg' box fusion because such combinations are shown the most effective in the original paper~\cite{chen2022multimodal}. Also, we adopt the E+M+T ensemble version, indicating that early-fusion, mid-fusion, and thermal single-modal detectors are ensembles.  \\
\indent \textbf{CMM}~\cite{kim2024causal}: The CMM framework is based on the Uncertainty-guided model~\cite{kim2021uncertainty}, with the FPN architecture with ResNet-50~\cite{he2016deep}. The learning rate is initialized at 0.007 during the initial 1 epoch and then a 0.1 learning rate decay is applied for the final epoch. The Region of Interests (RoIs) per image \ \ is set to N=300. We apply switchable Total Indirect Effect (sTIE) for every ROI and compute the prediction score. We used the official code, with the same settings as the original paper~\cite{kim2024causal}. \textbf{ICAFusion}: We used the official code provided by the authors. The model is trained for 60 epochs, with a batch size of 8. SGD~\cite{amari1993backpropagation} optimizer is used. The initial learning rate is $1.0 \times 10^{-2}$ with a momentum value of 0.937. The weight decay factor is 0.0005. The cosine annealing method is used for learning rate decay. \textbf{MS-DETR}: We used the code provided by the authors. The model is pretrained on the COCO\mbox{~\cite{lin2014microsoft}} dataset with the Resnet50~\cite{he2016deep} backbone. Transformer encoder and decoder consist of 6 layers. The number of feature scales, sampling points, object queries, and attention heads are $L = 4$, $K = 4$, $N = 300$, and $H = 8$, respectively. Adam~\cite{kingma2014adam} optimizer is used. The model is trained for 10 epochs with batch size 2. The initial learning rate is 0.0001 and is reduced by a factor of 10 halfway through the training process.
\section{Experimental Result}
\subsection{Result on FLIR, and CVC-14}
\begin{figure*}[t]
\centering
\includegraphics[width=0.9\textwidth]{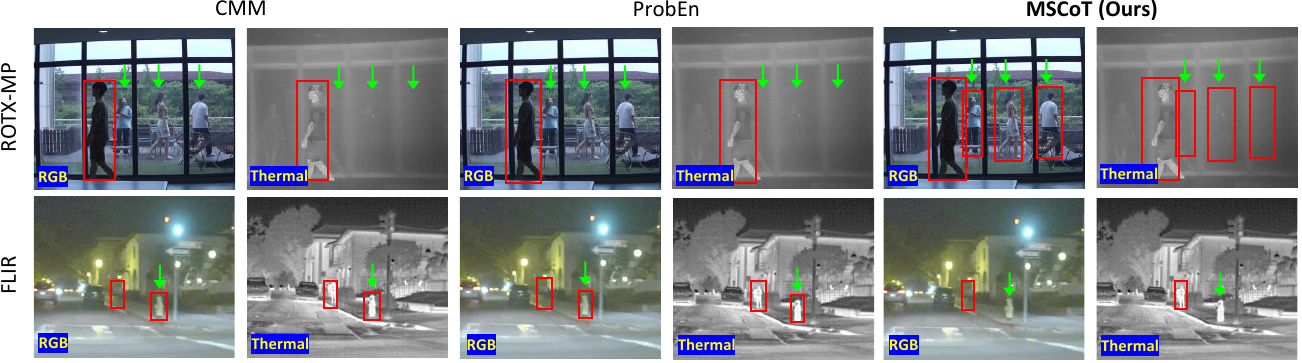}
\caption{Visualized detections on the ROTX-MP~\cite{kim2024causal} (Top) and FLIR~\cite{c:25} (Bottom) datasets. (Top): MSCoTDet can detect pedestrians over the window (\textcolor{green}{$\downarrow$}), in which their thermal signatures are absent due to the window. In contrast, other models (CMM~\cite{kim2024causal}, ProbEn~\cite{chen2022multimodal}) fail to detect these pedestrians. (Bottom): Moreover, other models create false-positive detections on the fireplug (\textcolor{green}{$\downarrow$}) due to its thermal signature similar-looking to pedestrians. In contrast, MSCoTDet (Ours) makes correct predictions.  }  
\end{figure*}
\begin{table*}[t!]
\caption{DETECTION PERFORMANCE ON THE FLIR~\cite{c:25}, CVC-14~\cite{gonzalez2016pedestrian} (LEFT), AND ROTX-MP~\cite{kim2024causal} (RIGHT). (LEFT): WE TRAIN AND TEST ON FLIR, CVC-14 DATASETS. (RIGHT): WE TRAIN MODELS ON FLIR, CVC-14, AND EVALUATE MODELS ON THE ROTX-MP DATASET. WE ADOPT THIS EXPERIMENTAL SETTING FOLLOWING THE ORIGINAL PAPER~\cite{kim2024causal} TO EVALUATE MODELS WHEN THERE IS A SIGNIFICANT DISTRIBUTION DIFFERENCE BETWEEN TRAIN AND TEST SETS. WE COMPARE OUR PROPOSED METHOD WITH DIFFERENT MULTISPECTRAL PEDESTRIAN DETECTION MODELS~\cite{liu2016multispectral, qingyun2022crossmodality, kim2021uncertainty, kim2024causal, chen2022multimodal, shen2024icafusion, xing2024ms} RECENTLY PROPOSED. THE BEST RESULTS ARE \textbf{HIGHLIGHTED}.}
	\renewcommand{\arraystretch}{1.4}
	\renewcommand{\tabcolsep}{1mm}
\centering
\resizebox{\linewidth}{!}{
\begin{tabular}{c|cccc|cccc|ccccc|cccc|cccc|ccc}
\clineB{1-12}{4} \clineB{14-25}{4}
Train                & \multicolumn{4}{c|}{FLIR}                                                             & \multicolumn{4}{c|}{CVC-14}                                                           & \multicolumn{3}{c}{\multirow{3}{*}{Modality Used}}                 &  & Train                & \multicolumn{4}{c|}{FLIR}                                                             & \multicolumn{4}{c|}{CVC-14}                                                           & \multicolumn{3}{c}{\multirow{3}{*}{Modality Used}}                 \\ 
\cline{1-9} \cline{14-22}
Test                 & \multicolumn{4}{c|}{FLIR}                                                             & \multicolumn{4}{c|}{CVC-14}                                                           & \multicolumn{3}{c}{}                                               &  & Test                 & \multicolumn{4}{c|}{ROTX-MP}                                                          & \multicolumn{4}{c|}{ROTX-MP}                                                          & \multicolumn{3}{c}{}                                               \\ 
\cline{1-9} \cline{14-22}
Metric               & \multicolumn{1}{c|}{AP($\uparrow$)}             & \multicolumn{3}{c|}{MR($\downarrow$)}                         & \multicolumn{1}{c|}{AP($\uparrow$)}             & \multicolumn{3}{c|}{MR($\downarrow$)}                         & \multicolumn{3}{c}{}                                               &  & Metric               & \multicolumn{1}{c|}{AP($\uparrow$)}             & \multicolumn{3}{c|}{MR($\downarrow$)}                         & \multicolumn{1}{c|}{AP($\uparrow$)}             & \multicolumn{3}{c|}{MR($\downarrow$)}                         & \multicolumn{3}{c}{}                                               \\ 
\cline{1-12} \cline{14-25} 
Method               & \multicolumn{1}{c|}{All}            & Day            & Night         & All            & \multicolumn{1}{c|}{All}            & Day           & Night          & All            & \multicolumn{1}{c|}{RGB} & \multicolumn{1}{c|}{Thermal} & Language &  & Method               & \multicolumn{1}{c|}{All}            & Day           & Night          & All            & \multicolumn{1}{c|}{All}            & Day           & Night          & All            & \multicolumn{1}{c|}{RGB} & \multicolumn{1}{c|}{Thermal} & Language \\ 
\cline{1-12} \cline{14-25} 
Halfway Fusion~\cite{liu2016multispectral}       & \multicolumn{1}{c|}{75.85}          & 42.13          & 36.75         & 40.65          & \multicolumn{1}{c|}{79.42}          & 36.29         & 26.29          & 31.99          & \multicolumn{1}{c|}{\checkmark}   & \multicolumn{1}{c|}{\checkmark}       &         &  & Halfway Fusion~\cite{liu2016multispectral}       & \multicolumn{1}{c|}{36.95}          & 69.76         & 51.45          & 59.13          & \multicolumn{1}{c|}{8.80}          & 86.65         & 94.36          & 92.01          & \multicolumn{1}{c|}{\checkmark}   & \multicolumn{1}{c|}{\checkmark}       &         \\
CFT~\cite{qingyun2022crossmodality}                  & \multicolumn{1}{c|}{84.10}          & 17.32          & 22.42         & 21.26          & \multicolumn{1}{c|}{88.45}          & 18.81         & 25.25          & 21.83          & \multicolumn{1}{c|}{\checkmark}   & \multicolumn{1}{c|}{\checkmark}       &         &  & CFT~\cite{qingyun2022crossmodality}                  & \multicolumn{1}{c|}{3.64}          & 81.51         & 87.68          & 83.83          & \multicolumn{1}{c|}{8.58}          & 98.01         & 89.89          & 94.38          & \multicolumn{1}{c|}{\checkmark}   & \multicolumn{1}{c|}{\checkmark}       &         \\
Kim et al.~\cite{kim2021uncertainty}           & \multicolumn{1}{c|}{84.67}          & 20.29          & 18.72         & 20.81          & \multicolumn{1}{c|}{90.08}          & 23.87         & 11.08          & 18.70          & \multicolumn{1}{c|}{\checkmark}   & \multicolumn{1}{c|}{\checkmark}       &         &  & Kim et al.~\cite{kim2021uncertainty}           & \multicolumn{1}{c|}{21.69}          & 63.99         & 60.68          & 64.02          & \multicolumn{1}{c|}{13.36}          & 73.54         & 98.75          & 84.49          & \multicolumn{1}{c|}{\checkmark}   & \multicolumn{1}{c|}{\checkmark}       &         \\
CMM~\cite{kim2024causal}                  & \multicolumn{1}{c|}{87.80}          & 17.81          & 13.01         & 16.60          & \multicolumn{1}{c|}{90.47}          & 27.81         & \textbf{7.71}           & 17.13          & \multicolumn{1}{c|}{\checkmark}   & \multicolumn{1}{c|}{\checkmark}       &         &  & CMM~\cite{kim2024causal}                  & \multicolumn{1}{c|}{57.09}          & 32.23         & 33.19           & 33.63          & \multicolumn{1}{c|}{34.96}          & 48.47         & 51.16           & 53.39          & \multicolumn{1}{c|}{\checkmark}   & \multicolumn{1}{c|}{\checkmark}       &         \\
ProbEn~\cite{chen2022multimodal}               & \multicolumn{1}{c|}{86.74}          & 21.29          & 10.16         & 17.45          & \multicolumn{1}{c|}{88.31}          & 23.01         & 21.02          & 23.23          & \multicolumn{1}{c|}{\checkmark}   & \multicolumn{1}{c|}{\checkmark}       &         &  & ProbEn~\cite{chen2022multimodal}               & \multicolumn{1}{c|}{17.20}          & 72.47         & 67.88          & 69.04          & \multicolumn{1}{c|}{16.66}          & 67.24         & 76.78          & 73.40          & \multicolumn{1}{c|}{\checkmark}   & \multicolumn{1}{c|}{\checkmark}       &         \\              
ICAFusion~\cite{shen2024icafusion}                  & \multicolumn{1}{c|}{86.80}          & 19.39          & 14.46         & 17.92          & \multicolumn{1}{c|}{88.43}          & 25.88         & 14.32           & 18.99          & \multicolumn{1}{c|}{\checkmark}   & \multicolumn{1}{c|}{\checkmark}       &         &  & ICAFusion~\cite{shen2024icafusion}                  & \multicolumn{1}{c|}{44.63}          & 38.58         & 59.92           & 47.10          & \multicolumn{1}{c|}{28.16}          & 53.63         & 63.18           & 56.63          & \multicolumn{1}{c|}{\checkmark}   & \multicolumn{1}{c|}{\checkmark}       &        \\ 
MS-DETR~\cite{xing2024ms}                   & \multicolumn{1}{c|}{88.08}          & 17.27          & 10.44         & 15.37          & \multicolumn{1}{c|}{90.79}          & 24.10         & 8.80           & 16.90          & \multicolumn{1}{c|}{\checkmark}   & \multicolumn{1}{c|}{\checkmark}       &         &  & 
MS-DETR~\cite{xing2024ms}                   & \multicolumn{1}{c|}{54.77}          & 27.96         & 40.95           & 35.33          & \multicolumn{1}{c|}{15.32}          & 75.42         & 75.70           & 76.24          & \multicolumn{1}{c|}{\checkmark}   & \multicolumn{1}{c|}{\checkmark}       &         \\ 
\cdashline{1-12} \cdashline{14-25} 
Halfway Fusion~\cite{liu2016multispectral} + LLM & \multicolumn{1}{c|}{69.82}          & 48.39          & 41.74         & 46.47          & \multicolumn{1}{c|}{78.14}          & 38.11         & 31.75          & 35.75          & \multicolumn{1}{c|}{\checkmark}   & \multicolumn{1}{c|}{\checkmark}       & \checkmark        &  & Halfway Fusion~\cite{liu2016multispectral} + LLM & \multicolumn{1}{c|}{32.88}          & 65.47         & 62.63          & 65.59          & \multicolumn{1}{c|}{6.84}          & 95.18         & 94.10          & 94.65          & \multicolumn{1}{c|}{\checkmark}   & \multicolumn{1}{c|}{\checkmark}       & \checkmark        \\
CFT~\cite{qingyun2022crossmodality} + LLM            & \multicolumn{1}{c|}{83.97}          & 16.65          & 26.54         & 23.91          & \multicolumn{1}{c|}{87.89}          & 17.01         & 30.12          & 23.81          & \multicolumn{1}{c|}{\checkmark}   & \multicolumn{1}{c|}{\checkmark}       & \checkmark        &  & CFT~\cite{qingyun2022crossmodality} + LLM            & \multicolumn{1}{c|}{0.95}          & 95.38         & 95.37          & 95.34          & \multicolumn{1}{c|}{2.90}          & 97.37         & 98.36          & 97.45          & \multicolumn{1}{c|}{\checkmark}   & \multicolumn{1}{c|}{\checkmark}       & \checkmark        \\
Kim et al.~\cite{kim2021uncertainty} + LLM     & \multicolumn{1}{c|}{84.26}          & 19.17          & 20.95         & 21.37          & \multicolumn{1}{c|}{89.47}          & 30.90         & 10.61          & 20.49          & \multicolumn{1}{c|}{\checkmark}   & \multicolumn{1}{c|}{\checkmark}       & \checkmark        &  & Kim et al.~\cite{kim2021uncertainty} + LLM     & \multicolumn{1}{c|}{19.17}          & 65.82         & 65.23          & 66.61          & \multicolumn{1}{c|}{12.65}          & 82.38         & 89.79          & 87.11          & \multicolumn{1}{c|}{\checkmark}   & \multicolumn{1}{c|}{\checkmark}       & \checkmark        \\
CMM~\cite{kim2024causal} + LLM            & \multicolumn{1}{c|}{87.79}          & 18.85          & 13.27         & 17.85          & \multicolumn{1}{c|}{91.39}          & 22.82         & 8.47           & 15.61          & \multicolumn{1}{c|}{\checkmark}   & \multicolumn{1}{c|}{\checkmark}       & \checkmark        &  & CMM~\cite{kim2024causal} + LLM            & \multicolumn{1}{c|}{37.39}          & 42.89         & 45.71           & 44.34          & \multicolumn{1}{c|}{37.93}           & 40.57         & 45.04           & 43.10          & \multicolumn{1}{c|}{\checkmark}   & \multicolumn{1}{c|}{\checkmark}       & \checkmark        \\
ProbEn~\cite{chen2022multimodal} + LLM         & \multicolumn{1}{c|}{89.38}          & 16.72          & 10.08         & 14.56          & \multicolumn{1}{c|}{89.50}          & 15.29         & 24.33          & 19.97          & \multicolumn{1}{c|}{\checkmark}   & \multicolumn{1}{c|}{\checkmark}       & \checkmark        &  & ProbEn~\cite{chen2022multimodal} + LLM         & \multicolumn{1}{c|}{24.55}          & 61.38         & 64.09          & 62.75          & \multicolumn{1}{c|}{23.56}          & 62.17         & 53.65          & 59.45          & \multicolumn{1}{c|}{\checkmark}   & \multicolumn{1}{c|}{\checkmark}       & \checkmark        \\     
ICAFusion~\cite{shen2024icafusion} + LLM                  & \multicolumn{1}{c|}{86.01}          & 20.46          & 14.45         & 18.56          & \multicolumn{1}{c|}{88.51}          & 24.95         & 13.21           & 18.38          & \multicolumn{1}{c|}{\checkmark}   & \multicolumn{1}{c|}{\checkmark}       & \checkmark        &  & ICAFusion~\cite{shen2024icafusion} + LLM                  & \multicolumn{1}{c|}{44.00}          & 39.75         & 61.16           & 48.12          & \multicolumn{1}{c|}{21.78}          & 55.36         & 67.81           & 65.37          & \multicolumn{1}{c|}{\checkmark}   & \multicolumn{1}{c|}{\checkmark}       & \checkmark         \\ 
MS-DETR~\cite{xing2024ms} + LLM                  & \multicolumn{1}{c|}{87.64}          & 17.61          & 10.89         & 15.68          & \multicolumn{1}{c|}{90.64}          & 26.12         & 9.27           & 18.30          & \multicolumn{1}{c|}{\checkmark}   & \multicolumn{1}{c|}{\checkmark}       & \checkmark        &  & MS-DETR~\cite{xing2024ms} + LLM                  & \multicolumn{1}{c|}{30.28}          & 43.49         & 54.23           & 55.24          & \multicolumn{1}{c|}{9.74}          & 90.64         & 89.77           & 90.11          & \multicolumn{1}{c|}{\checkmark}   & \multicolumn{1}{c|}{\checkmark}       & \checkmark         \\ 
\cdashline{1-12} \cdashline{14-25} 
MSCoTDet(Ours)       & \multicolumn{1}{c|}{\textbf{90.39}} & \textbf{13.99} & \textbf{8.38} & \textbf{12.16} & \multicolumn{1}{c|}{\textbf{92.38}} & \textbf{6.69} & 13.69 & \textbf{10.39} & \multicolumn{1}{c|}{\checkmark}   & \multicolumn{1}{c|}{\checkmark}       & \checkmark        &  & MSCoTDet(Ours)       & \multicolumn{1}{c|}{\textbf{67.33}} & \textbf{15.23} & \textbf{24.02} & \textbf{23.57} & \multicolumn{1}{c|}{\textbf{57.65}} & \textbf{27.16} & \textbf{34.35} & \textbf{34.18} & \multicolumn{1}{c|}{\checkmark}   & \multicolumn{1}{c|}{\checkmark}       & \checkmark        \\ 
\clineB{1-12}{4} \clineB{14-25}{4} 
\end{tabular}}
\label{table1}
\end{table*}

We train and test on FLIR~\cite{c:25}, CVC-14~\cite{gonzalez2016pedestrian} datasets to evaluate whether our method performs well on general multispectral pedestrian data. We report the detection performance of different models on these datasets. For the evaluation metrics, we use average precision (AP) and log-average miss rate (MR). Following previous works\mbox{~\cite{kim2021uncertainty,kim2024causal,qingyun2022crossmodality,chen2022multimodal,liu2016multispectral}}, we report MR performance separately for daytime (‘Day’), nighttime (‘Night’), and the entire set (‘All’), while AP is evaluated on the entire set (‘All’). Note that lower MR (MR $\downarrow$) and higher AP (AP $\uparrow$) values indicate better detection performance. \\
\indent Table \ref{table1} (Left) shows the experimental results on the FLIR and CVC-14 test set. On FLIR, MSCoTDet achieves the highest AP (90.39 AP), and the lowest MR (13.99, 8.38, and 12.16 MR). Such results indicate the effectiveness of MSCoTDet on well-aligned data. On CVC-14, MSCoTDet demonstrates the lowest miss rate among models in (`Day'), and (`All'), with a value of 6.69 MR and 10.39 MR, respectively. Also, the highest AP (92.38 AP) is achieved. As (`All') indicates the entire dataset, our method achieves the best performance also on CVC-14. Such results illustrate that MSCoTDet is also effective on a dataset that largely contains misaligned multispectral data.
\begin{table*}[t!]
\caption{ABLATION STUDY ON THE EFFECT OF USING LANGUAGE MODELS. OUR FOCUS IS TO COMPARE MSCOTDET WITH ‘LATE-FUSION’, THE VISION BRANCH EQUIPPED WITH VISION-DRIVEN DETECTION OF MSCOTDET.}
	\renewcommand{\arraystretch}{1.4}
	\renewcommand{\tabcolsep}{1mm}
\centering
\resizebox{\linewidth}{!}{
\begin{tabular}{c|cccc|cccc|ccccc|cccc|cccc|ccc}
\clineB{1-12}{4} \clineB{14-25}{4}
Train          & \multicolumn{4}{c|}{FLIR}                            & \multicolumn{4}{c|}{CVC-14}                          & \multicolumn{3}{c}{\multirow{3}{*}{Modality Used}}                 &  & Train          & \multicolumn{4}{c|}{FLIR}                            & \multicolumn{4}{c|}{CVC-14}                          & \multicolumn{3}{c}{\multirow{3}{*}{Modality Used}}                 \\ 
\cline{1-9} \cline{14-22}
Test           & \multicolumn{4}{c|}{FLIR}                            & \multicolumn{4}{c|}{CVC-14}                          & \multicolumn{3}{c}{}                                               &  & Test           & \multicolumn{4}{c|}{ROTX-MP}                         & \multicolumn{4}{c|}{ROTX-MP}                         & \multicolumn{3}{c}{}                                               \\ 
\cline{1-9} \cline{14-22}
Metric         & \multicolumn{1}{c|}{AP($\uparrow$)}    & \multicolumn{3}{c|}{MR($\downarrow$)} & \multicolumn{1}{c|}{AP($\uparrow$)}    & \multicolumn{3}{c|}{MR($\downarrow$)} & \multicolumn{3}{c}{}                                               &  & Metric         & \multicolumn{1}{c|}{AP($\uparrow$)}    & \multicolumn{3}{c|}{MR($\downarrow$)} & \multicolumn{1}{c|}{AP($\uparrow$)}    & \multicolumn{3}{c|}{MR($\downarrow$)} & \multicolumn{3}{c}{}                                               \\ 
\cline{1-12} \cline{14-25} 
Method         & \multicolumn{1}{c|}{All}   & Day    & Night  & All   & \multicolumn{1}{c|}{All}   & Day    & Night  & All   & \multicolumn{1}{c|}{RGB} & \multicolumn{1}{c|}{Thermal} & Language &  & Method         & \multicolumn{1}{c|}{All}   & Day    & Night  & All   & \multicolumn{1}{c|}{All}   & Day    & Night  & All   & \multicolumn{1}{c|}{RGB} & \multicolumn{1}{c|}{Thermal} & Language \\ 
\cline{1-12} \cline{14-25} 
RGB only       & \multicolumn{1}{c|}{76.06} & 23.23  & 33.18  & 26.33 & \multicolumn{1}{c|}{86.54} & 31.55  & 40.10  & 36.07 & \multicolumn{1}{c|}{\checkmark}   & \multicolumn{1}{c|}{}        &          &  & RGB only       & \multicolumn{1}{c|}{54.20} & 32.47  & 36.22  & 35.61 & \multicolumn{1}{c|}{31.05} & 30.82  & 55.39  & 52.21 & \multicolumn{1}{c|}{\checkmark}   & \multicolumn{1}{c|}{}        &          \\
Thermal only   & \multicolumn{1}{c|}{85.73} & 21.56  & 15.84  & 20.17 & \multicolumn{1}{c|}{79.72} & 43.24  & 33.44  & 36.07 & \multicolumn{1}{c|}{}    & \multicolumn{1}{c|}{\checkmark}       &          &  & Thermal only   & \multicolumn{1}{c|}{19.33} & 72.47  & 67.88  & 69.04 & \multicolumn{1}{c|}{9.10}  & 91.61  & 92.41  & 91.97 & \multicolumn{1}{c|}{}    & \multicolumn{1}{c|}{\checkmark}       &          \\
Late-fusion    & \multicolumn{1}{c|}{88.60} & 17.31  & 11.68  & 16.14 & \multicolumn{1}{c|}{88.53} & 18.33  & 18.86  & 21.57 & \multicolumn{1}{c|}{\checkmark}   & \multicolumn{1}{c|}{\checkmark}       &          &  & Late-fusion    & \multicolumn{1}{c|}{58.95} & 27.42  & 34.89  & 32.67 & \multicolumn{1}{c|}{44.72} & 38.82  & 44.78  & 42.20 & \multicolumn{1}{c|}{\checkmark}   & \multicolumn{1}{c|}{\checkmark}       &          \\ 
\cdashline{1-12} \cdashline{14-25} 
MSCoTDet(Ours) & \multicolumn{1}{c|}{\textbf{90.39}} & \textbf{13.99} & \textbf{8.38} & \textbf{12.16} & \multicolumn{1}{c|}{\textbf{92.38}} & \textbf{6.69} & \textbf{13.69} & \textbf{10.39} & \multicolumn{1}{c|}{\checkmark}   & \multicolumn{1}{c|}{\checkmark}       & \checkmark        &  & MSCoTDet(Ours) & \multicolumn{1}{c|}{\textbf{67.33}} & \textbf{15.23} & \textbf{24.02} & \textbf{23.57} & \multicolumn{1}{c|}{\textbf{57.65}} & \textbf{27.16} & \textbf{34.35} & \textbf{34.18} & \multicolumn{1}{c|}{\checkmark}   & \multicolumn{1}{c|}{\checkmark}       & \checkmark        \\ 
\clineB{1-12}{4} \clineB{14-25}{4}
\end{tabular}}
\label{language}
\end{table*}
\begin{table*}[t!]
\caption{ABLATION STUDY ON THE SCORE-FUSION STRATEGY. WE COMPARE THE 1) AVG, AVG, 2) AVG, MAX, 3) MAX, MAX, AND 4) MAX, AVG STRATEGIES IN THE VISION-DRIVEN DETECTION (‘V’) AND IN THE VISION-LANGUAGE FUSION (‘VL’).}
	\renewcommand{\arraystretch}{1.3}
	\renewcommand{\tabcolsep}{1.0mm}
\centering
\resizebox{0.7\linewidth}{!}{
\begin{tabular}{cc|cccc|ccccccc|cccc|cccc}
\clineB{1-10}{4} \clineB{12-21}{4}
\multicolumn{2}{c|}{Train}     & \multicolumn{4}{c|}{FLIR}                            & \multicolumn{4}{c}{CVC-14}                          &  & \multicolumn{2}{c|}{Train}     & \multicolumn{4}{c|}{FLIR}                            & \multicolumn{4}{c}{CVC-14}                          \\ \cline{1-10} \cline{12-21} 
\multicolumn{2}{c|}{Test}      & \multicolumn{4}{c|}{FLIR}                            & \multicolumn{4}{c}{CVC-14}                          &  & \multicolumn{2}{c|}{Test}      & \multicolumn{4}{c|}{ROTX-MP}                            & \multicolumn{4}{c}{ROTX-MP}                          \\ \cline{1-10} \cline{12-21} 
\multicolumn{2}{c|}{Metric}    & \multicolumn{1}{c|}{AP($\uparrow$)}    & \multicolumn{3}{c|}{MR($\downarrow$)} & \multicolumn{1}{c|}{AP($\uparrow$)}    & \multicolumn{3}{c}{MR($\downarrow$)} &  & \multicolumn{2}{c|}{Metric}    & \multicolumn{1}{c|}{AP($\uparrow$)}    & \multicolumn{3}{c|}{MR($\downarrow$)} & \multicolumn{1}{c|}{AP($\uparrow$)}    & \multicolumn{3}{c}{MR($\downarrow$)} \\ 
\cline{1-10} \cline{12-21}
\multicolumn{1}{c|}{V}   & VL  & \multicolumn{1}{c|}{All}   & Day    & Night  & All   & \multicolumn{1}{c|}{All} & Day &Night   & All   &  & \multicolumn{1}{c|}{V}   & VL  & \multicolumn{1}{c|}{All}   & Day    & Night  & All   & \multicolumn{1}{c|}{All}   & Day    & Night & All   \\ 
\cline{1-11} \cline{12-21}  
\multicolumn{1}{c|}{Avg} & Avg & \multicolumn{1}{c|}{89.47} & 15.87  & 9.94   & 14.20 & \multicolumn{1}{c|}{89.68} & 10.98  & 17.38 & 16.12 &  & \multicolumn{1}{c|}{Avg} & Avg & \multicolumn{1}{c|}{66.22} & 20.14  & 27.20  & 26.92 & \multicolumn{1}{c|}{56.69} & 35.56  & 38.90 & 36.14 \\
\multicolumn{1}{c|}{Avg} & Max & \multicolumn{1}{c|}{88.31} & 16.71  & 10.76  & 15.34 & \multicolumn{1}{c|}{88.86} & 10.26  & 18.53 & 15.71 &  & \multicolumn{1}{c|}{Avg} & Max & \multicolumn{1}{c|}{66.27} & 28.17  & 26.89  & 27.67 & \multicolumn{1}{c|}{56.60} & 34.80  & 37.19 & 36.69 \\
\multicolumn{1}{c|}{Max} & Max & \multicolumn{1}{c|}{89.94} & 16.89  & 9.03   & 14.57 & \multicolumn{1}{c|}{89.64} & 8.83   & 17.54 & 14.07 &  & \multicolumn{1}{c|}{Max} & Max & \multicolumn{1}{c|}{66.31} & 26.14  & 25.29  & 25.78 & \multicolumn{1}{c|}{56.02} & 31.43  & 39.38 & 37.12 \\
\multicolumn{1}{c|}{Max} & Avg & \multicolumn{1}{c|}{\textbf{90.39}} & \textbf{13.99}  & \textbf{8.38}   & \textbf{12.16} & \multicolumn{1}{c|}{\textbf{92.38}} & \textbf{6.69}   & \textbf{13.69} & \textbf{10.39} &  & \multicolumn{1}{c|}{Max} & Avg & \multicolumn{1}{c|}{\textbf{67.33}} & \textbf{15.23}  & \textbf{24.02}  & \textbf{23.57} & \multicolumn{1}{c|}{\textbf{57.65}} & \textbf{27.16}  & \textbf{34.35} & \textbf{34.18} \\ 
\clineB{1-10}{4} \clineB{12-21}{4}
\end{tabular}}
\label{ablation-score}
\end{table*}
\vspace{-0.3cm}
\subsection{Result on ROTX-MP}
Experimenting with models on the ROTX-MP~\cite{kim2024causal} dataset is to evaluate modality bias and their generalizability when there is a significant distributional change in test data. Following the original work~\cite{kim2024causal}, we train models on general datasets (FLIR~\cite{c:25}, and CVC-14~\cite{gonzalez2016pedestrian}) and test models on ROTX-MP. The results are in Table \ref{table1} (Right). MSCoTDet achieves the best performance on ROTX-MP for both cases when trained on FLIR and CVC-14. When trained from FLIR, MSCoTDet achieves the highest AP (67.33 AP) and lowest MR (15.23, 24.02, 23.57 MR) which outperforms other methods by at least 10.24 AP, 27.66 MR, 9.17 MR, and 10.06 MR, respectively. When trained from CVC-14, MSCoTDet achieves the highest AP (57.65 AP) and lowest MR (27.16, 34.35, 34.18) which outperforms other methods by at least 22.69 AP, 13.41 MR, 10.69 MR, and 8.92 MR, respectively. Such results indicate that MSCoTDet has a better ability to intervene in modality bias compared to other models. 
\subsection{Comparison with Competing Methods that Provided Access to an LLM}
\indent For a fair comparison, we also evaluate the competing methods~\cite{kim2024causal,kim2021uncertainty,qingyun2022crossmodality, liu2016multispectral, chen2022multimodal, shen2024icafusion, xing2024ms} that provided access to using a pre-trained LLM~\cite{chatgpt, achiam2023gpt}. As with MSCoTDet, text descriptions of RGB/thermal pedestrians are generated with an MLLM~\cite{achiam2023gpt}. The same MSCoTDet implementations are adopted for competing methods of late-fusion~\cite{chen2022multimodal}. For competing methods of mid-fusion\mbox{~\cite{kim2024causal,kim2021uncertainty,qingyun2022crossmodality, liu2016multispectral, shen2024icafusion, xing2024ms}}, we modify MSCoT prompting and LMF to ensure compatibility. The modifications involved using the same bounding box regions for RGB and thermal images in MSCoT Prompting, (i.e., $B_{RGB}^{V}=B_{T}^{V}=B_{F}^{V}$ in Eq.11) since mid-fusion models produce only fused outputs. Accordingly, the LMF is modified to substitute $B_{F}^{V}$ to $B_{F}^{L}$ in Eq.12. Other processes are kept the same, and the evaluation is performed with the detection results obtained through Eq.12. \\
\indent The results of the Halfway Fusion\mbox{~\cite{liu2016multispectral}} + LLM, CFT\mbox{~\cite{qingyun2022crossmodality}}+LLM, Kim et al.\mbox{~\cite{kim2021uncertainty}} + LLM, CMM\mbox{~\cite{kim2024causal}} + LLM, ProbEn\mbox{~\cite{chen2022multimodal}} + LLM, ICAFusion\mbox{~\cite{shen2024icafusion}} + LLM, \mbox{MS-DETR~\cite{xing2024ms}} + LLM on the FLIR~\cite{c:25} dataset, CVC-14\mbox{~\cite{gonzalez2016pedestrian}}, and ROTX-MP\mbox{~\cite{kim2024causal}} are shown in TABLE I. MSCoTDet consistently outperforms these competing methods on all datasets as other methods show only marginal improvements or even degrade with LLM integration. These results suggest that MSCoTDet effectively incorporates the LLM into the detection process, and thereby enhances detection accuracy.
\section{Ablation Study}
\subsection{Effect of Integrating Language Models}
We conduct an ablation study to evaluate the effectiveness of integrating large language models to our framework. As our framework fuses detection results from both the vision branch and the language branch, one might ask: \textit{how much does the language branch boost overall performance?} For the ablation study, we compare MSCoTDet and three detection models: (1) `RGB only', (2) `Thermal only', and (3) `Late-fusion'. First, the `RGB only', and the `Thermal only' model indicate single-modal detection in the vision branch of MSCoTDet. Second, `Late-fusion' is in which the single-modal detections of `RGB only' and `Thermal only' are fused, i.e., equivalent to the vision branch equipped with vision-driven detection of our framework. Thus, to measure the effect of using large language models, our main focus is to compare `Late-fusion' with MSCoTDet. The results are shown in Table \ref{language}. Our method achieves higher performance than `Late-fusion' in all FLIR~\cite{c:25}, CVC-14~\cite{gonzalez2016pedestrian}, and ROTX-MP~\cite{kim2024causal}. The results on the FLIR and CVC-14 indicate that integrating large language models can improve the overall performance of multispectral pedestrian detectors in general datasets. Also, the results on ROTX-MP indicate that using language models can significantly improve the generalizability of multispectral pedestrian detectors, and effectively mitigate the modality bias.

\begin{table*}[t!]
\caption{ABLATION STUDY ON THE BOX-FUSION STRATEGY. WE EXPERIMENT ARGMAX AND S-AVG FOR EACH VISION-DRIVEN DETECTION (‘V’), LANGUAGE-DRIVEN DETECTION (‘L’), AND VISION-LANGUAGE FUSION (‘VL’).}
	\renewcommand{\arraystretch}{1.3}
	\renewcommand{\tabcolsep}{1.0mm}
\centering
\resizebox{0.9\linewidth}{!}{
\begin{tabular}{ccc|cccc|cccccccc|cccc|cccc}
\clineB{1-11}{4} \clineB{13-23}{4}
\multicolumn{3}{c|}{Train}                                         & \multicolumn{4}{c|}{FLIR}                            & \multicolumn{4}{c}{CVC-14}                          &  & \multicolumn{3}{c|}{Train}                                         & \multicolumn{4}{c|}{FLIR}                            & \multicolumn{4}{c}{CVC-14}                          \\ 
\cline{1-11} \cline{13-23} 
\multicolumn{3}{c|}{Test}                                          & \multicolumn{4}{c|}{FLIR}                            & \multicolumn{4}{c}{CVC-14}                          &  & \multicolumn{3}{c|}{Test}                                          & \multicolumn{4}{c|}{ROTX-MP}                         & \multicolumn{4}{c}{ROTX-MP}                         \\ 
\cline{1-11} \cline{13-23} 
\multicolumn{3}{c|}{Method}                                        & \multicolumn{1}{c|}{AP($\uparrow$)}    & \multicolumn{3}{c|}{MR($\downarrow$)} & \multicolumn{1}{c|}{AP($\uparrow$)}    & \multicolumn{3}{c}{MR($\downarrow$)} &  & \multicolumn{3}{c|}{Method}                                        & \multicolumn{1}{c|}{AP($\uparrow$)}    & \multicolumn{3}{c|}{MR($\downarrow$)} & \multicolumn{1}{c|}{AP($\uparrow$)}    & \multicolumn{3}{c}{MR($\downarrow$)} \\ 
\cline{1-11} \cline{13-23}
\multicolumn{1}{c|}{V}      & \multicolumn{1}{c|}{L}      & VL     & \multicolumn{1}{c|}{All}   & Day    & Night  & All   & \multicolumn{1}{c|}{All}   & Day    & Night & All   &  & \multicolumn{1}{c|}{V}      & \multicolumn{1}{c|}{L}      & VL     & \multicolumn{1}{c|}{All}   & Day    & Night  & All   & \multicolumn{1}{c|}{All}   & Day    & Night & All   \\ 
\cline{1-11} \cline{13-23} 
\multicolumn{1}{c|}{argmax} & \multicolumn{1}{c|}{argmax} & argmax & \multicolumn{1}{c|}{89.49} & 15.31  & 9.43   & 13.65 & \multicolumn{1}{c|}{91.06} & 7.56   & 17.64 & 14.00 &  & \multicolumn{1}{c|}{argmax} & \multicolumn{1}{c|}{argmax} & argmax & \multicolumn{1}{c|}{66.10} & 15.35  & 31.88  & 26.16 & \multicolumn{1}{c|}{56.97} & 35.94  & 29.39 & 35.61 \\
\multicolumn{1}{c|}{s-avg}  & \multicolumn{1}{c|}{argmax} & argmax & \multicolumn{1}{c|}{89.79} & 14.59  & 8.82   & 12.75 & \multicolumn{1}{c|}{91.60} & 8.03  & 15.28 & 12.63 &  & \multicolumn{1}{c|}{s-avg}  & \multicolumn{1}{c|}{argmax} & argmax & \multicolumn{1}{c|}{66.29} & 18.64  & 25.95  & 25.86 & \multicolumn{1}{c|}{56.95} & 29.89  & 37.75 & 36.17 \\
\multicolumn{1}{c|}{argmax} & \multicolumn{1}{c|}{s-avg}  & argmax & \multicolumn{1}{c|}{89.83} & 14.88  & 8.97   & 13.12 & \multicolumn{1}{c|}{90.86} & 8.32   & 15.39 & 13.31 &  & \multicolumn{1}{c|}{argmax} & \multicolumn{1}{c|}{s-avg}  & argmax & \multicolumn{1}{c|}{67.03} & 26.16  & \textbf{23.30}  & 25.07 & \multicolumn{1}{c|}{56.90} & 35.44  & 29.38 & 35.07 \\
\multicolumn{1}{c|}{s-avg}  & \multicolumn{1}{c|}{s-avg}  & argmax & \multicolumn{1}{c|}{90.32} & 14.22  & 8.77   & 12.47 & \multicolumn{1}{c|}{91.72} & 7.60   & 15.14 & 12.54 &  & \multicolumn{1}{c|}{s-avg}  & \multicolumn{1}{c|}{s-avg}  & argmax & \multicolumn{1}{c|}{67.24} & 19.51  & 25.95  & 26.14 & \multicolumn{1}{c|}{57.13} & 29.88  & 37.27 & 36.07 \\
\multicolumn{1}{c|}{argmax} & \multicolumn{1}{c|}{argmax} & s-avg  & \multicolumn{1}{c|}{89.49} & 15.07  & 8.73   & 13.18 & \multicolumn{1}{c|}{90.71} & \textbf{6.25}   & 18.51 & 14.18 &  & \multicolumn{1}{c|}{argmax} & \multicolumn{1}{c|}{argmax} & s-avg  & \multicolumn{1}{c|}{66.11} & 24.76  & 23.67  & 26.57 & \multicolumn{1}{c|}{57.01} & 35.07  & 29.27 & 35.23 \\
\multicolumn{1}{c|}{s-avg}  & \multicolumn{1}{c|}{argmax} & s-avg  & \multicolumn{1}{c|}{90.32} & 14.25  & 8.55   & 12.43 & \multicolumn{1}{c|}{92.12} & 7.63   & 15.19 & 11.99 &  & \multicolumn{1}{c|}{s-avg}  & \multicolumn{1}{c|}{argmax} & s-avg  & \multicolumn{1}{c|}{67.28} & 17.53  & 25.95  & 25.60 & \multicolumn{1}{c|}{56.99} & 29.97  & 37.75 & 36.22 \\
\multicolumn{1}{c|}{argmax} & \multicolumn{1}{c|}{s-avg}  & s-avg  & \multicolumn{1}{c|}{89.73} & 14.13  & 9.26   & 12.76 & \multicolumn{1}{c|}{91.41} & 6.85   & 16.81 & 13.47 &  & \multicolumn{1}{c|}{argmax} & \multicolumn{1}{c|}{s-avg}  & s-avg  & \multicolumn{1}{c|}{67.15} & 25.08  & 23.67  & 26.69 & \multicolumn{1}{c|}{57.21} & 35.34  & \textbf{29.26} & 35.07 \\
\multicolumn{1}{c|}{s-avg}  & \multicolumn{1}{c|}{s-avg}  & s-avg  & \multicolumn{1}{c|}{\textbf{90.39}} & \textbf{13.99}  & \textbf{8.38}   & \textbf{12.16} & \multicolumn{1}{c|}{\textbf{92.38}} & 6.69   & \textbf{13.69} & \textbf{10.39} &  & \multicolumn{1}{c|}{s-avg}  & \multicolumn{1}{c|}{s-avg}  & s-avg  & \multicolumn{1}{c|}{\textbf{67.33}} & \textbf{15.23}  & 24.02  & \textbf{23.57} & \multicolumn{1}{c|}{\textbf{57.65}} & \textbf{27.16}  & 34.35 & \textbf{34.18} \\ 
\clineB{1-11}{4} \clineB{13-23}{4}
\end{tabular}}
\label{ablation-box}
\end{table*}

\begin{table*}[t!]
\caption{ABLATION STUDY OF USING DIFFERENT MLLM MODELS FOR GENERATING TEXT DESCRIPTIONS. FOUR DIFFERENT MODELS (PHANTOM-7B\cite{lee2024phantom}, INTERNVL-8B\cite{chen2024internvl}, GEMINI-1.5 PRO\cite{team2023gemini}, GPT-4V\cite{achiam2023gpt}) ARE USED FOR THE EXPERIMENTS. EXPERIMENTS ARE CONDUCTED ON FLIR, CVC-14 AND ROTX-MP DATASETS.}
	\renewcommand{\arraystretch}{1.3}
	\renewcommand{\tabcolsep}{1.0mm}
\centering
\resizebox{0.8\linewidth}{!}{
\begin{tabular}{cc|cccc|ccccccc|cccc|cccc}
\clineB{1-10}{4} \clineB{12-21}{4}
\multicolumn{2}{c|}{Train}     & \multicolumn{4}{c|}{FLIR}                           & \multicolumn{4}{c}{CVC-14}                          &  & \multicolumn{2}{c|}{Train}     & \multicolumn{4}{c|}{FLIR}                            & \multicolumn{4}{c}{CVC-14}                          \\ 
\cline{1-10} \cline{12-21} 
\multicolumn{2}{c|}{Test}      & \multicolumn{4}{c|}{FLIR}                            & \multicolumn{4}{c}{CVC-14}                          &  & \multicolumn{2}{c|}{Test}      & \multicolumn{4}{c|}{FLIR}                            & \multicolumn{4}{c}{CVC-14}                          \\ 
\cline{1-10} \cline{12-21} 
\multicolumn{2}{c|}{Metric}    & \multicolumn{1}{c|}{AP($\uparrow$)}    & \multicolumn{3}{c|}{MR($\downarrow$)} & \multicolumn{1}{c|}{AP($\uparrow$)}    & \multicolumn{3}{c}{MR($\downarrow$)} &  & \multicolumn{2}{c|}{Method}    & \multicolumn{1}{c|}{AP($\uparrow$)}    & \multicolumn{3}{c|}{MR($\downarrow$)} & \multicolumn{1}{c|}{AP($\uparrow$)}    & \multicolumn{3}{c}{MR($\downarrow$)} \\ 
\cline{1-10} \cline{12-21} 
\multicolumn{2}{c|}{Metric}  & \multicolumn{1}{c|}{All}   & Day    & Night  & All   & \multicolumn{1}{c|}{All}   & Day    & Night & All   &  & \multicolumn{2}{c|}{Method}    & \multicolumn{1}{c|}{All}   & Day    & Night  & All   & \multicolumn{1}{c|}{All}   & Day    & Night & All   \\ 
\cline{1-11} \cline{12-21} 
\multicolumn{2}{c|}{Phantom-7B\cite{lee2024phantom}} & \multicolumn{1}{c|}{90.08} & 16.31  & 8.88   & 13.84 & \multicolumn{1}{c|}{91.51} & 8.46  & 14.19 & 11.32 &  & \multicolumn{2}{c|}{Phantom-7B\cite{lee2024phantom}} & \multicolumn{1}{c|}{66.22} & 23.28  & 28.92  & 24.73 & \multicolumn{1}{c|}{56.43} & 28.82  & 34.89 & 33.38 \\
\multicolumn{2}{c|}{InternVL-8B\cite{chen2024internvl}} & \multicolumn{1}{c|}{90.18} & 15.73  & 9.18  & 13.78 & \multicolumn{1}{c|}{91.48} & 8.56  & 14.75 & 11.29 &  & \multicolumn{2}{c|}{InternVL-8B\cite{chen2024internvl}} & \multicolumn{1}{c|}{66.32} & 24.09  & 27.46  & 24.67 & \multicolumn{1}{c|}{56.84} & 24.27  & 34.79 & 33.51 \\
\multicolumn{2}{c|}{Gemini-1.5 Pro\cite{team2023gemini}} & \multicolumn{1}{c|}{90.31} & 16.29  & \textbf{7.62}   & 13.53 & \multicolumn{1}{c|}{92.02} & 9.51   & 14.51 & 11.46 &  & \multicolumn{2}{c|}{Gemini-1.5 Pro\cite{team2023gemini}} & \multicolumn{1}{c|}{66.94} & 23.79  & 27.41  & 24.38 & \multicolumn{1}{c|}{57.09} & \textbf{23.68}  & 34.57 & \textbf{32.53} \\
\multicolumn{2}{c|}{GPT-4V\cite{achiam2023gpt}} & \multicolumn{1}{c|}{\textbf{90.39}} & \textbf{13.99}  & 8.38   & \textbf{12.16} & \multicolumn{1}{c|}{\textbf{92.38}} & \textbf{6.69}   & \textbf{13.69} & \textbf{10.39} &  & \multicolumn{2}{c|}{GPT-4V\cite{achiam2023gpt}} & \multicolumn{1}{c|}{\textbf{67.33}} & \textbf{15.23}  & \textbf{24.02}  & \textbf{23.57} & \multicolumn{1}{c|}{\textbf{57.65}} & 27.16  & \textbf{34.35} & 34.18 \\ 
\clineB{1-10}{4} \clineB{12-21}{4}
\end{tabular}}
\end{table*}

\subsection{Effect of the Fusion Strategy}
\label{sec:6.2}
We conduct an ablation study to evaluate the design choice of our fusion strategy in Section \ref{sec:3.4}. Our fusion strategies include score-fusion and box-fusion. For score-fusion, we adopt the Non-maximum Suppression (NMS) in the vision-driven detection (`V') and averaging strategy in the vision-language fusion (`VL'), respectively.  NMS compares the prediction scores estimated in each modality and votes for the highest one, removing the lower scores (Eq. \ref{eq:2}). The averaging strategy averages prediction scores estimated in each modality (Eq. \ref{eq:1}). Denote the averaging strategy as `Avg' and NMS as `Max'. For the comparison study, we conducted experiments on the (1) Avg, Avg, (2) Avg, Max, and (3) Max, Max fusion strategies in each vision-driven detection (`V') and vision-language fusion (`VL'), respectively. We do not compare different score-fusion in the language branch, as we use our proposed method: MSCoT (Section \ref{sec:3.3}) produces prediction scores from fused modalities. All other conditions are kept the same. The results are shown in Table \ref{ablation-score}. The Max, Avg strategy consistently performs the best in all test sets. \\
\indent For box-fusion, we adopt weighted averaging (Eq. \ref{eq:3}) for all vision-driven detection (`V'), language-driven detection (`L'), and vision-language fusion (`VL'). Denote the weighted averaging strategy as `s-avg'. For the comparison study, we consider the NMS box fusion, denoted as `argmax', and compare between a total of 8 combinations (either s-avg or max in three parts). All other conditions are kept the same as our MSCoTDet design during all experiments. The results are shown in Table \ref{ablation-box}. Applying s-avg for all V, L, and VL shows the best performance for most test cases.
\subsection{Impact of Using Different MLLM Models for Generating Text Descriptions}
We conduct an ablation study to evaluate the impact of using different MLLM models for generating text descriptions. MSCoTDet is tested with four different MLLMs: Phantom-7B~\cite{lee2024phantom}, InternVL-8B~\cite{chen2024internvl}, Gemini-1.5 Pro~\cite{team2023gemini} and GPT-4V~\cite{achiam2023gpt}. Throughout these experiments, other components of MSCoTDet remained the same.\\
\indent The results in Table V show that MSCoTDet maintains consistent performance across the various MLLMs used for text generation, with the best results obtained with GPT-4V. This indicates that MSCoTDet remains robust regardless of the specific MLLM used, making it adaptable and effective across different MLLMs.

\begin{table}[t!]
\caption{Computational comparison on the model efficiency regarding the inclusion of the language model. The average running time (seconds per image) of MSCoTDet with and without the language branch is evaluated on the FLIR~\cite{c:25} dataset.}
\renewcommand{\arraystretch}{1.1}
\renewcommand{\tabcolsep}{2.0mm}
\centering
\resizebox{0.9\linewidth}{!}{
\begin{tabular}{c|c|c}
\clineB{1-3}{4}
\multicolumn{1}{c|}{\multirow{2}{*}{Method}} & \multicolumn{1}{c|}{\multirow{2}{*}{Running Time (s)}} & Use of the \\ 
&&Language Model\\ \cline{1-3}
\multicolumn{1}{c|}{MSCoTDet without language branch} & \multicolumn{1}{c|}{0.16}   &           \\ \cdashline{1-3}
\multicolumn{1}{c|}{MSCoTDet}         & \multicolumn{1}{c|}{0.20}   & \checkmark           \\ \clineB{1-3}{4}
\end{tabular}}
\end{table}
\vspace{-0.2cm}
\subsection{Impact of the Inclusion of the Language Model on the Model's Efficiency}
\indent To quantitatively analyze how the inclusion of the language model affects the model’s efficiency, we compare MSCoTDet with and without the language branch. The MSCoTDet without the language branch is equivalent to the visual branch of the MSCoTDet alone. The computational comparison was conducted by measuring the model’s average running time (seconds per image) on the FLIR~\cite{c:25} dataset, using a single Nvidia A6000 GPU. The results reported in Table VI show that the running time of MSCoTDet with and without the language branch is 0.20 seconds and 0.16 seconds per image, respectively. Thus, the inclusion of the language model adds an additional 0.04 seconds per image to the running time. 
\section{Discussion}
\indent \textbf{Limitation and Future Work.} The goal of our work is to address modality bias in multispectral pedestrian detection and enhance detection accuracy using large language models (LLMs). A limitation is that our work focuses on a single class: pedestrians. As language models can generate descriptions and detection scores for classes other than pedestrians, our Multispectral Chain-of-Thought (MSCoT) prompting method can potentially be integrated into object detection on general target classes. Moreover, Language-Driven Multi-modal Fusion (LMF) is applicable to fuse the detection results of various classes by conducting class-wise operations in eq.10-eq.12. We plan to extend our work to other object classes in future work.\\
\indent \textbf{Applicability to Other Multi-modality Tasks.} Our proposed methods can be potentially applied in other multi-modality fusion tasks in computer vision. For instance, our strategies of integrating textual descriptions across different modalities can help to train models when there are insufficient annotations, e.g., as in scribble-supervised RGB-D saliency detection~\cite{li2023robust}. Another potential application is in the medical field, where multiple imaging modalities (e.g., T1-weighted, T1-contrast, T2-weighted, and FLAIR MRIs) are used in combination to detect/segment lesions (e.g., brain tumor segmentation~\cite{zhang2021cross, zhang2020exploring}). Our technique of generating detection rationales from multi-modal images can be applied to generate clinical evidence when detecting lesions from multiple medical images.
\vspace{-0.3cm}
\section{Conclusion}
In this paper, we introduced MSCoTDet, a novel framework for enhancing multispectral pedestrian detection by addressing modality bias using a Large Language Model (LLM). The framework employs a Multispectral Chain-of-Thought (MSCoT) prompting strategy to guide an LLM in performing multispectral detection tasks, integrating this with vision-based models through a Language-driven Multi-modal Fusion (LMF) approach. Our extensive experiments demonstrated that MSCoTDet outperforms existing methods on standard datasets such as FLIR and CVC-14, achieving significant improvements in Average Precision (AP) and Miss Rates. Furthermore, MSCoTDet achieved impressive performance on the ROTX-MP dataset, proving the generalizability, particularly in scenarios with distribution shifts between training and testing data.
\\


\vspace{-0.3cm}
\footnotesize
\bibliographystyle{IEEEbib}
{\linespread{0.9}\selectfont\bibliography{refs}}
\vspace{-4em}
\begin{IEEEbiography}[\raisebox{0.35in}{\includegraphics[width=0.8in,height=1in,clip]{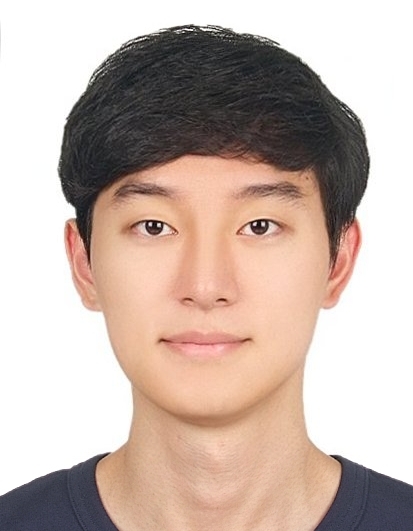}}]{TAEHEON KIM}
received the B.S. degree in electrical engineering from Korea Advanced Institute of Science and Technology (KAIST), Daejeon, South Korea in 2019. He is currently pursuing the Ph.D. in electrical engineering at KAIST, Daejeon, South Korea. His research interests include deep learning, multimodal learning, causal inference, object detection, and adversarial robustness.
\end{IEEEbiography}
\vspace{-6em}
\begin{IEEEbiography}[\raisebox{0.35in}{\includegraphics[width=0.8in,height=1in,clip]{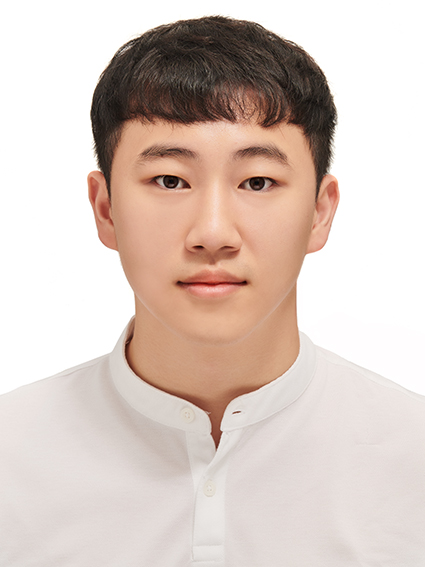}}]{SANGYUN CHUNG}
 received the B.S. degree in computer science from Hanyang University, Seoul, South Korea in 2023. He is currently pursuing the M.S. in electrical engineering at KAIST, Daejeon, South Korea. His research interests include deep learning, multimodal learning, object detection, large language models, and adversarial robustness.
\end{IEEEbiography}
\vspace{-6em}
\begin{IEEEbiography}[\raisebox{0.45in}{\includegraphics[width=0.8in,height=1in,clip]{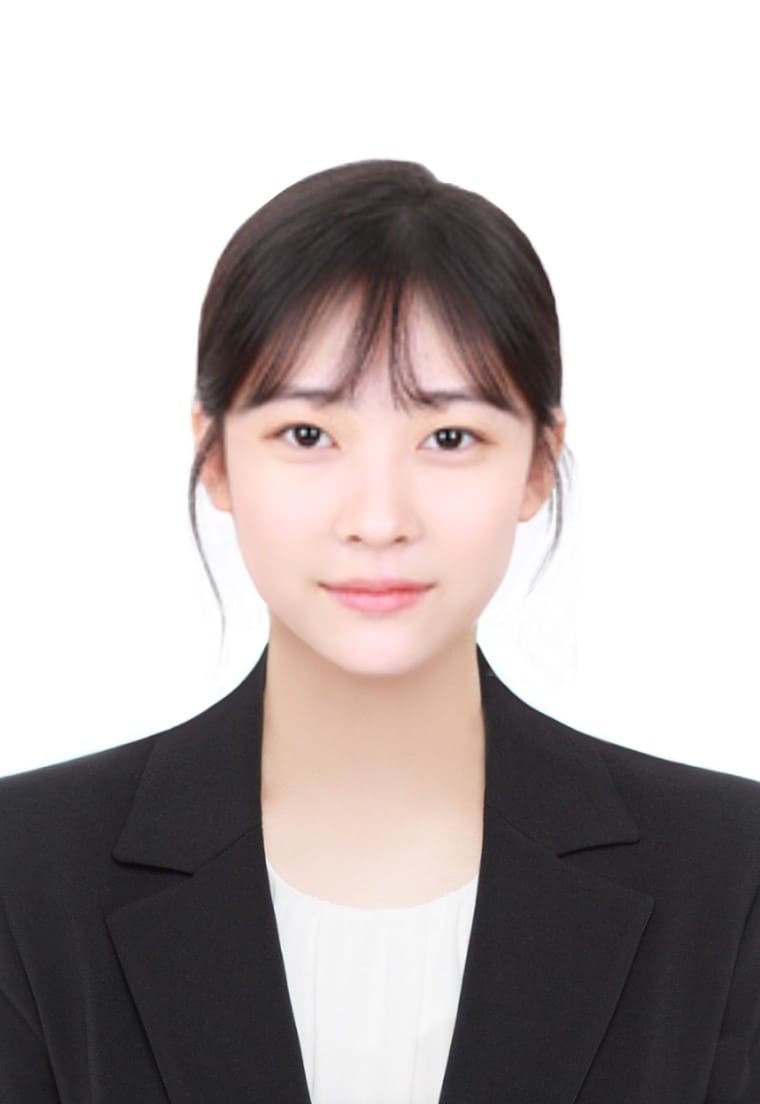}}]{DAMIN YEOM}
 received the B.S. degree in electronic and electrical engineering from Ewha Womans University, Seoul, South Korea in 2023. She is currently pursuing the M.S. in electrical engineering at KAIST, Daejeon, South Korea. Her research interests include deep learning, object detection, multi-modal learning.
\end{IEEEbiography}
\vspace{-6em}
\begin{IEEEbiography}[\raisebox{0.2in}{\includegraphics[width=0.8in,height=1in,clip]{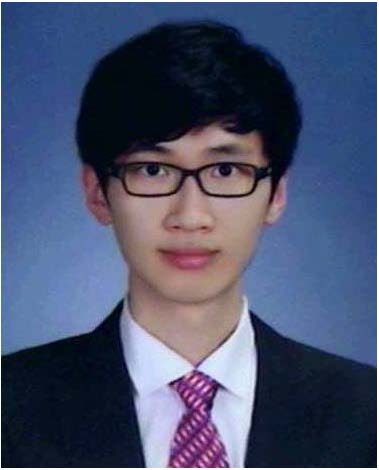}}]{YOUNGJOON YU}
received the B.S. degree in electrical engineering from Korea Advanced Institute of Science and Technology (KAIST), Daejeon, South Korea in 2013, and the M.S. degree in the management engineering from KAIST in 2017. He is currently pursuing the Ph.D. in electrical engineering at KAIST, Daejeon, South Korea. His research interests include deep learning, multi-sensor learning, and adversarial robustness.
\end{IEEEbiography}
\vspace{-5em}
\begin{IEEEbiography}[
{\includegraphics[width=0.8in,height=1in,clip]{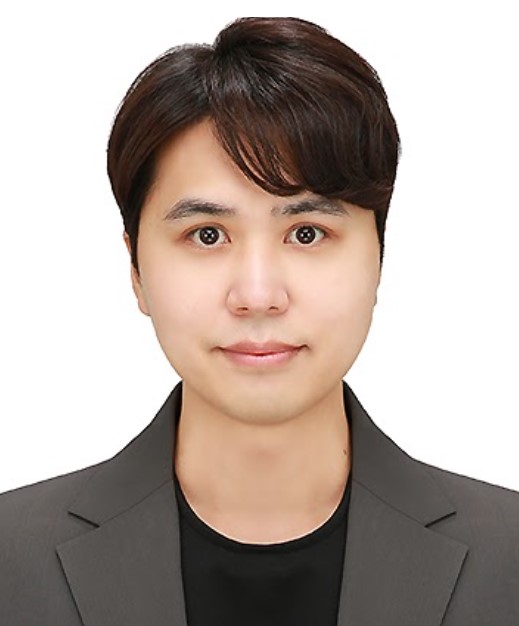}}]{HAK GU KIM}
received the B.S. and M.S. degrees from Inha University, Incheon, South Korea, in 2012 and 2014, respectively, and the Ph.D. degree from the Korea Advanced Institute of Science and Technology (KAIST), Daejeon, South Korea, in 2019. He was a Postdoctoral Researcher with École Polytechnique Fédérale de Lausanne (EPFL), Lausanne, Switzerland. He is currently an Assistant Professor with the Graduate School of Advanced Imaging Science, Multimedia \& Films (GSAIM), Chung-Ang University, Seoul, South Korea. His research interests include deep learning and machine learning in 2D/3D/VR image and video processing and computer vision, human visual perception, and multi-modal learning.
\end{IEEEbiography}
\begin{IEEEbiography}[
{\includegraphics[width=0.9in,height=1.125in,clip]{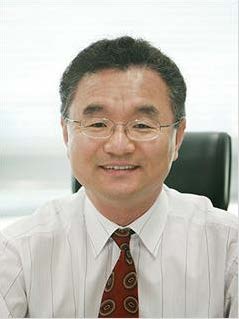}}]{YONG MAN RO}
 (Senior Member, IEEE) received his Ph.D. degree from the School of Electrical Engineering at KAIST. He conducted research at various institutions, including Columbia University, the University of California, Irvine, and the Department of Electrical Engineering and Computer Sciences at the University of California, Berkeley. He also served as a visiting professor at The Edward S. Rogers Sr. Department of Electrical and Computer Engineering at the University of Toronto.
He is currently ICT endowed chair professor at the School of Electrical Engineering at KAIST. He is also the director of the Center for Applied Research in Artificial Intelligence (CARAI), the Image Video System Lab, and the Integrated Vision and Language Lab at KAIST. Throughout his career, he has conducted research on a wide range of image and video system topics. His research interests include image processing, computer vision, multimodal deep learning, integrating vision, speech, and language for AI, multimodal object and motion detection/recognition, inclusive human multimodal conversation, and analysis for interpretability and robustness of deep learning models. He received the young investigator finalist award from ISMRM in 1992 and the Scientist of the Year Award (South Korea) in 2003. He has served as a TPC member for numerous international conferences, including roles as program chair and organizer of special sessions. He was an Associate Editor for IEEE Signal Processing Letters and currently serves as an Associate Editor for IEEE Transactions on Circuits and Systems for Video Technology.
\end{IEEEbiography}

\end{document}